\documentclass{article}

 \usepackage[preprint]{neurips_2026}

\usepackage[utf8]{inputenc} 
\usepackage[T1]{fontenc}    
\usepackage{url}            
\usepackage{booktabs}       
\usepackage{amsfonts}       
\usepackage{nicefrac}       
\usepackage{microtype}      
\usepackage{xcolor}         

\usepackage[dvipsnames]{xcolor}
\usepackage[colorlinks=true,linkcolor=MidnightBlue,citecolor=gray, urlcolor=gray]{hyperref}

\usepackage{enumitem}
\setlist[itemize]{topsep=0em}

\usepackage{amsmath}
\usepackage{pifont}
\usepackage{url}
\usepackage[nameinlink]{cleveref}
\crefname{equation}{Eq.}{Eqs.} 
\Crefname{equation}{Eq.}{Eqs.}
\crefname{algocf}{Alg.}{Algs.}
\crefname{section}{Sec.}{Secs.}
\crefname{figure}{Fig.}{Figs.}
\crefname{appsection}{App.}{Apps.}

\usepackage[ruled,vlined]{algorithm2e}
\usepackage{bbm}
\usepackage{graphicx}

\usepackage{inconsolata}

\usepackage{newtxtext,newtxmath}
\usepackage[frak=esstix]{mathalfa}

\usepackage{subcaption}
\usepackage{wrapfig}
\usepackage{lipsum}
\usepackage{booktabs}
\usepackage{makecell}
\usepackage{xspace}
\usepackage{enumitem}
\usepackage{multirow}

\newcommand{\acronym}{SpeedAug\xspace}

\title{SpeedAug: Policy Acceleration via Tempo-Enriched Policy and RL Fine-Tuning}

\newcommand{\kaist}[0]{\textsuperscript{1}}
\newcommand{\unist}[0]{\textsuperscript{2}}
\newcommand{\deepauto}[0]{\textsuperscript{3}}

\author{%
  Taewook Nam\kaist, Junmo Cho\kaist, Youngsoo Jang\unist, Sung Ju Hwang\kaist\textsuperscript{,}\deepauto \\
\kaist KAIST, \unist UNIST, \deepauto DeepAuto.ai\\
\texttt{\{namsan, sjhwang82\}@kaist.ac.kr}
}

\begin{document}
    \maketitle

    \begin{abstract}
Robotic policy learning for complex real-world manipulation tasks has seen rapid recent progress, enabled in large part by the ability to collect demonstrations through human operation.
However, policies trained from such demonstrations often execute tasks far more slowly than the robot’s physical capabilities, as demonstration data is collected under practical constraints that favor conservative, success-oriented trajectories over execution speed.
Existing policy acceleration methods determine execution tempo through data preprocessing or heuristic rules, rather than learning execution speed optimized for the task.
In this paper, we propose SpeedAug, a policy acceleration framework that enables policies to learn task-optimal execution tempo via reinforcement learning (RL).
SpeedAug first learns a tempo-enriched prior policy from speed-augmented demonstrations that captures diverse execution tempos.
Building on this tempo-enriched prior, RL fine-tuning guides exploration to refine action trajectories and optimize execution tempo efficiently.
Experiments on robotic manipulation benchmarks demonstrate that SpeedAug substantially improves the sample efficiency of policy acceleration while maintaining high success rates, achieving fast and stable task execution. Applied to a real-world manipulation task, SpeedAug improves task throughput by 1.8× using only 16 minutes of online interactions without compromising the success rate.
\end{abstract}
    \section{Introduction}
Robotic policy learning has made rapid progress in recent years, driven by scalable data collection through human teleoperation and advances in expressive policy representations such as diffusion-based models~\citep{chi2024diffusionpolicy,mobilealoha,realtimeactionchunking,bjorck2025gr00t,pi0.5}.
However, existing studies primarily focus on solving challenging tasks with a feasible amount of data, resulting in policies whose execution speeds often fall short of the hardware’s capabilities~\citep{guo2025demospeedup,yuan2025speedtuning,arachchige2025sail,park2024proleptic,kim2025espada}.
Collecting demonstrations that execute tasks faster typically requires more data and higher demonstration proficiency, which is costly in domains like robot control where large-scale data collection is already expensive.

A line of research aims to accelerate task execution by skipping intermediate actions~\citep{park2024proleptic,guo2025demospeedup,yuan2025speedtuning,kim2025espada} or by executing actions at a higher control frequency~\citep{arachchige2025sail}.
To avoid overspeeding task segments that require precise, slower motions, these works selectively accelerate actions depending on the task phase.
Yet, synthetically accelerated actions and hard decisions on when to accelerate inevitably introduce distribution shifts from the original demonstration execution speeds, thereby motivating further policy adaptation.

A promising approach is to fine-tune pre-trained policies via reinforcement learning (RL).
With sparse rewards, RL can further improve task success rates without requiring dense human feedback.
However, standard RL relies on unguided random exploration after policy pre-training, which does not explicitly encourage exploration at varying action speeds.
In contrast, once humans learn how to complete a task, they deliberately perform the same behavior at different tempos and gradually refine their skills to achieve the fastest feasible execution.

Inspired by this observation, we propose \texttt{\textbf{\acronym}}, an RL-based policy acceleration framework designed to efficiently explore and acquire faster task execution behaviors.
\texttt{\acronym} fine-tunes a pre-trained policy using sparse rewards and a standard RL algorithm to reduce task completion time while preserving the high success rate of the pre-trained policy.
To mimic the human strategy of intentionally testing different tempos of action, we pre-train the policy on speed-augmented demonstrations that expose it to a range of action tempos, including unseen tempos faster than the original demonstrations.
Starting from this tempo-enriched policy, the subsequent RL fine-tuning jointly selects fast yet safe tempos and refines the accelerated actions, further improving task execution speed and the success rate of the accelerated behavior.

To obtain the tempo-enriched policy, we pre-train a generative imitation learning policy on speed-augmented demonstration data.
The speed-augmented data is generated by synthetically accelerating the actions in the original demonstrations using acceleration factors sampled from a predefined distribution.
Specifically, each action sequence in the demonstrations is downsampled with a randomly sampled factor to synthesize accelerated actions, while deferring the decision of feasible action speed and action refinement to the fine-tuning phase.
Behavior cloning with action chunking~\citep{aloha,chi2024diffusionpolicy} embeds the arbitrary tempos in the augmented action sequences into the policy, leveraging the diffusion-based policy's multimodal action distribution.

We evaluate the effectiveness of \texttt{\acronym} across simulated manipulation benchmarks and a real robot task.
In simulation benchmarks, \texttt{\acronym} consistently achieves 2–3× faster execution while maintaining over 99\% success rates, with fewer online samples and fewer failures than baseline methods.
Applied to a continual pick-and-place task on robot 
hardware, \texttt{\acronym} improves task throughput by 1.8× using only 16 minutes of 
real-world experience, without compromising the success rate.
Our analyses on synthetic 
navigation and manipulation tasks provide insights into the performance gains of 
\texttt{\acronym} and the suboptimal learning efficiency of baseline methods.



    \begin{figure*}[t]
    \centering
    \includegraphics[width=0.98\linewidth]{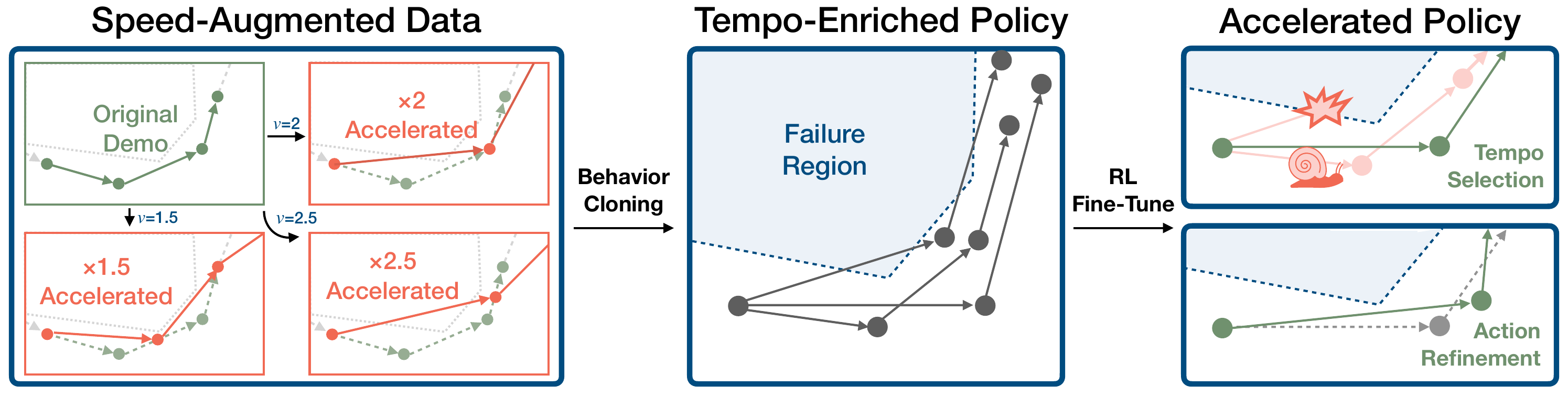}

    \caption{\textbf{Our Framework.} We propose to fine-tune tempo-enriched policy by RL for efficient policy acceleration. Our pre-training with speed-augmentation of demonstrations construct a behavior prior that embeds diverse tempos of action accelerations. Subsequent RL fine-tuning jointly explores feasible tempos and adapt the accelerated actions in online environment, quickly converging to faster behavior with high success rate.}
\vspace{-0.4cm}
\label{fig:method}
\end{figure*}

\section{Preliminaries}
 
\subsection{Problem Formulation}
We define the task as a Markov Decision Process (MDP), denoted by $\mathcal{M} = \langle \mathcal{S}, \mathcal{A}, \mathcal{P}, R, \gamma \rangle$, where $\mathcal{S}$ is the state space, $\mathcal{A}$ is the action space, $\mathcal{P}$ is the transition dynamics, $R$ is the reward function, and $\gamma$ is the discount factor.
The reward is sparse and provided only upon task completion, i.e., $r_t = 1$ if the task is completed and $r_t = 0$ otherwise.

We assume access to demonstration data $\mathcal{D}=\{ (s_{1:T_i}, a_{1:T_i})\}_{i=1\dots N}$, consisting of successful demonstrations typically collected by human operators via teleoperations~\citep{aloha,chi2024diffusionpolicy,mobilealoha,khazatsky2024droid}.
Since the primary goal of robot learning is to achieve a high success rate, $S(\pi) = \mathbb{E}_\pi[\sum_t r_t]$, the quantity and quality of the collected demonstrations are often sufficient to train an imitation learning policy $\pi$ with high $S(\pi)$.
However, task execution time, $E(\pi) = \mathbb{E}_\pi[t|r_t=1]$, is typically less prioritized than success rate due to constraints such as data collection cost, teleoperation interfaces, or demonstrator proficiency~\citep{arachchige2025sail,guo2025demospeedup,yuan2025speedtuning}.
Therefore, we assume that a policy $\pi_\mathcal{D}$ pre-trained on $\mathcal{D}$ completes the task reliably but substantially slower than the optimal policy $\pi^*$, i.e., $S(\pi_\mathcal{D}) \approx S(\pi^*)$ and $E(\pi_\mathcal{D}) > E(\pi^*)$.

Our goal is to obtain a faster yet stable policy $\pi$ from $\mathcal{D}$, i.e., $S(\pi) \approx S(\pi^*)$ and $E(\pi) \approx E(\pi^*)$.
Standard RL methods can address this objective by fine-tuning $\pi_\mathcal{D}$ with a time-penalty reward or by simply maximizing the discounted reward, $\sum_{t=1}^T \gamma^{T-t}r_t$, with $\gamma < 1$.
However, in practice, optimizing execution time via a discounted reward is highly sample-inefficient when the pre-trained policy’s action distribution is concentrated around slow tempos, forcing RL to rely on unguided exploration.
We evaluate the performance of standard RL and demonstrate that our method improves sample efficiency without compromising the final task performance.

\subsection{Imitation Learning for Robotics}
\textbf{Generative Policy.}
Following common design choices in recent robotics policies~\citep{pi0.5,liu2024rdt,bjorck2025gr00t}, we model the policy as a generative model with an iterative process to obtain a high performance pre-trained policy from demonstrations.
Specifically, we adopt Diffusion Policy~\citep{chi2024diffusionpolicy} for our simulation experiments, in which action sampling is performed via the denoising process of diffusion probabilistic models (DDPM) 
\begin{equation}
    \mathbf{a}^{(K)} \sim \mathcal{N}(0, \mathbf{I}),~~~
    \mathbf{a}^{(k-1)} \sim \mathcal{N} \left(\mu_\theta(\mathbf{s}, \mathbf{a}^{(k)}) , \tilde{\beta}_k \mathbf{I} \right) ,~~~\mathbf{a} = \mathbf{a}^{(0)},
\label{eq:ddpm}
\end{equation}

given the number of noising steps $K$,  noise schedule $\beta_k$, $\alpha_k=1-\beta_k$, $\bar{\alpha}_k=\prod_{i=1}^{k}\alpha_i$,  $\mu_\theta(\mathbf{s}, \mathbf{a}^{(k)}) =\frac{1}{\sqrt{\alpha_k}} ( \mathbf{a}^{(k)} -  \frac{1-\alpha_k}{\sqrt{1 - \bar{\alpha}_k}} \epsilon_\theta(\mathbf{a}^{(k)}, \mathbf{s}, k))$, and $ \tilde{\beta}_k=\frac{1-\bar{\alpha}_{k-1}}{1-\bar{\alpha}_k}\beta_k$.
The denoising network $\epsilon_\theta$ is trained by minimizing
\begin{equation}
    \mathbb{E}_{\mathbf{s}, \mathbf{a} \sim \mathcal{D},k\sim U(1, K)} \left[
      \| \epsilon^{(k)} - \epsilon_\theta(\mathbf{a}^{(k)}, \mathbf{s}, k) \|^2
    \right].
\label{eq:ddpm_loss}
\end{equation}
\citet{chi2024diffusionpolicy} show that the diffusion-based architecture captures multimodal action distributions in high-dimensional action spaces.
We leverage this capacity to embed a variety of action tempos into the policy output distribution.
Our real robot experiments use a flow matching~\citep{lipman2022flow} policy, 
which replaces the SDE in~\Cref{eq:ddpm} with a deterministic ODE, sharing comparable generative capacity with DDPM with fewer iterations.

\textbf{Action-chunking.}
Another key design choice is action chunking~\citep{chi2024diffusionpolicy,aloha}, which predicts a sequence of actions for a given state input.
In this scheme, a single policy sampling produces $H$ or more actions at once, which are then executed sequentially over $H$ steps before the next sampling.
This can be implemented by sampling subsequences of states and actions, $(\mathbf{s}_t, \mathbf{a}_t) = (s_{t:t+H-1}, a_{t:t+H-1})$, from the demonstrations, where non-bold notation denotes the actual environment state and action.
While the primary benefit of action chunking is to enhance temporal consistency in imitation learning and support high-frequency control, we further utilize it to enable the policy to predict the action tempo over a near-future horizon.

\subsection{Policy Acceleration and Tempo Selection}
Pre-trained policies often execute tasks conservatively, resulting in slow task execution to achieve high success rate~\citep{guo2025demospeedup,yuan2025speedtuning,arachchige2025sail,park2024proleptic,kim2025espada}.
A common approach to policy acceleration assumes position-based control, where a policy predicts a sequence of absolute target configurations, such as end-effector poses or joint configurations~\citep{chi2023diffusionpolicy,guo2025demospeedup,arachchige2025sail,yuan2025speedtuning}.
The predicted target configurations are usually processed by a low-level controller to produce the final control signals.
In this setting, acceleration can be achieved by downsampling the action sequence in demonstration data~\citep{guo2025demospeedup} or in the predicted action sequence~\citep{yuan2025speedtuning}, or by executing the original sequence at a higher control frequency~\citep{arachchige2025sail} through appropriate controller parameterization.

However, naively accelerating actions across the entire course of a task degrades the success rate, since certain task segments require slower and more precise motion.
Previous works have proposed state-conditioned action acceleration based on heuristic decisions~\citep{guo2025demospeedup,arachchige2025sail,kim2025espada} or learned RL policies~\citep{yuan2025speedtuning} to prevent failure by selectively accelerating actions while avoiding acceleration in precision-critical segments.
Throughout this paper, we refer to whether and at what level an action segment is accelerated as its \emph{tempo}, since it determines how fast the given action sequence is interpreted, distinguishing it from more general terms such as action speed or acceleration.

    \section{Approach}

\subsection{Motivation}
While recent works have demonstrated that conditionally adjusting action \emph{tempo} can effectively accelerate task execution while retaining success rates~\citep{guo2025demospeedup,arachchige2025sail,kim2025espada}, synthetically accelerated actions are deployed in a real environment without further modification, based on hard decisions made at the pre-training stage.
Because downsampled actions skip intermediate states both temporally and spatially, the resulting behaviors can oversimplify trajectories or cause overshooting by ignoring low-level controller dynamics (e.g., accelerating too aggressively and requiring abrupt stopping).
Executing original action sequence at an increased control frequency with a high-gain controller can reduce this distributional gap, but it requires increased action computation throughput and adjustment of low-level controller parameters~\citep{arachchige2025sail}.
Alternatively, conservatively selecting faster tempos mitigates these issues but often leads to suboptimal execution speed.
These observations naturally motivate adapting synthetically accelerated actions within the actual execution environment.

RL fine-tuning of an accelerated policy is a direct yet unexplored approach to address this motivation.
Since the accelerated policy already attains a non-zero success rate, RL can in principle adapt the policy to further improve both success rate and execution time using only sparse reward feedback.
However, once a policy is pre-trained with fixed tempo selections, subsequent exploration during RL fine-tuning relies on unguided random exploration, without access to arbitrary action tempos.
As a result, if the initial tempo decisions are suboptimal, overcoming this limitation can be highly sample-inefficient.
Moreover, as the policy is fine-tuned, the optimal tempo decisions may change dynamically alongside the progress of action adaptation.
For example, aggressively accelerating a task segment may degrade success rates before fine-tuning, but after sufficient adaptation, the same acceleration can become successful while significantly reducing execution time.
Training a separate tempo-selection policy, as suggested in~\citet{yuan2025speedtuning}, is a promising approach, but introduces additional technical challenges in integrating tempo selection with fine-tuning the pre-trained policy.
Thus, we seek to devise a method that simultaneously enables diverse tempo exploration and effective action adaptation.

To this end, we propose \texttt{\acronym}, an RL-based framework that efficiently accelerates the task completion speed of a pre-trained policy.
We first construct a \emph{tempo-enriched policy} that embeds diverse action tempos (\Cref{sec:relabel}), and then perform RL fine-tuning that jointly selects safe tempos and refines actions through standard online RL policy updates (\Cref{sec:rl}).


\subsection{Tempo-Enriched Policy Pre-training}\label{sec:relabel}

To enable exploration of diverse tempos jointly with policy fine-tuning, we propose embedding various action tempos into the initial policy used for online RL.
Starting from this \emph{tempo-enriched policy}, subsequent RL fine-tuning implicitly performs tempo selection while simultaneously adapting the policy to the actual execution environment.
To obtain such a policy, we pre-train it on demonstration data $\mathcal{D}$ using accelerated action chunks with randomly sampled tempos.
To formalize this process, we first define an acceleration function by downsampling a given position-based action sequence with tempo $v$:
\begin{equation}
    \mathrm{Accel}(\mathbf{a}_t, v) = [ a_{t+v-1}, a_{t+2v-1}, \cdots, a_{t+(H-1)v-1}],
\label{eq:accel}
\end{equation}
where it can generalize to non-integer action index using linear interpolation: $a_{vt}=a_{\lfloor vt \rfloor} +  \frac{vt - \lfloor vt \rfloor}{v}(a_{\lceil vt \rceil }-a_{\lfloor vt \rfloor })$.
Then, we train a diffusion-based policy using randomly accelerated action target:
\begin{equation}
    \mathbb{E}_{\substack{\mathbf{s},\mathbf{a} \sim \mathcal{D},k\sim U(1,K),\\v\sim U(1,v_\text{max})}}
    \left[
      \| \epsilon_k - \epsilon_\theta(\text{Accel}(\mathbf{a}, v)+\epsilon_k, \mathbf{s}, k) \|^2
    \right]
\label{eq:ddpm_loss}
\end{equation}
where $U(a, b)$ is a uniform distribution over $[a,b]$, and $v_\text{max}$ controls the maximum tempo.
The resulting policy covers a range of accelerated action tempos, leveraging the multimodal action distribution of the generative policy and action chunking scheme.
Although naively executing this policy may lead to high variance in task completion, it serves as an initial policy for subsequent RL fine-tuning, promoting efficient exploration and refining it into a stable and fast policy.

\subsection{RL Fine-Tuning}\label{sec:rl}
The pre-trained tempo-enriched policy is used as an initialization for RL fine-tuning.
The benefits of using our proposed pre-trained policy are twofold.
First, diverse action tempos are now in-distribution samples from the policy’s action distribution, allowing the policy to structurally explore and gradually select feasible tempos.
This is distinguished from prior works that explicitly and rigidly select action tempo during pre-training~\citep{guo2025demospeedup,arachchige2025sail,kim2025espada}, making our tempo selection implicit.
Second, even when the accelerated behavior is suboptimal due to synthetic acceleration via downsampling (e.g., a slight drop in success rate), subsequent fine-tuning leveraging the retained diversity of tempos can refine these imperfect behaviors into stable actions.

We perform RL fine-tuning using sparse rewards and a discount factor $\gamma < 1$.
Maximizing the discounted return effectively optimizes the execution time of the pre-trained policy without compromising its success rate.
DPPO~\citep{dppo2024} and PA-RL~\citep{parl} are used as the RL algorithms for our experiments, while any other RL algorithms that are compatible with diffusion- or flow-based policies can be used.
    \section{Experiments}

In our experiments, we demonstrate that \texttt{\acronym} can accelerate task execution speed of a policy without compromising success rate, using fewer environment interactions compared to baseline methods.
We first summarize considered baseline methods (\Cref{sec:baselines}), then validate the methods on simulated manipulation benchmarks (\Cref{sec:main_result}) and a real robot task (\Cref{sec:robot_exp}).
For a more intuitive illustration of our method, 
experiments on a maze navigation task are also provided 
in~\Cref{app:synthetic}.

\subsection{Baselines}\label{sec:baselines}
The following baselines are considered throughout our experiments:
\begin{itemize}[leftmargin=1.3em]
\itemsep0.4em
\parskip0em
\parsep0em
\item \texttt{No-Accel}: RL fine-tuning of the pre-trained policy with original non-accelerated demonstrations.
\item \texttt{Accel($v$)}: pre-training with demonstrations accelerated by a constant factor $v$ using~\Cref{eq:accel}, followed by RL fine-tuning.
\item \texttt{SpeedTuning}~\citep{yuan2025speedtuning}: Training a separate policy to decide the tempos conditioned on the current state while keeping the pre-trained policy frozen.
\item \texttt{DemoSpeedup(+RL)~\citep{guo2025demospeedup}}: Pre-training with partially accelerated demonstrations based on tempos determined by action entropy of a proxy policy, followed by RL fine-tuning.
\end{itemize}
See~\Cref{app:baseline_detail} for implementation details.

\begin{figure*}[t]
    \centering
    \includegraphics[width=\linewidth]{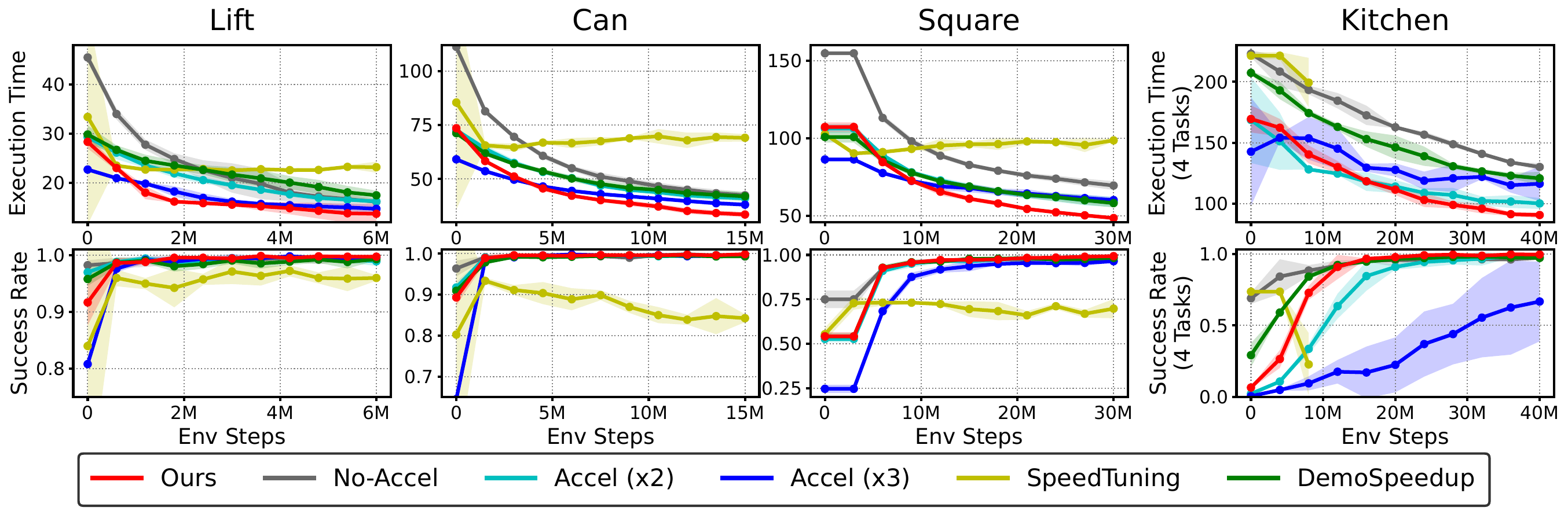}
    \caption{\textbf{Manipulation Benchmark Results.} Learning progress of success rate and task completion time over environment steps collected. \texttt{\acronym} achieves faster execution using fewer online samples while maintaining high success rates, compared to other baselines.}
\label{fig:main_result}
\vspace{-0.2cm}
\end{figure*}
\newcommand{\mc}[1]{\multicolumn{2}{c}{#1}}
\newcommand{\mcr}[1]{\multicolumn{2}{c|}{#1}}
\newcommand{\q}[1]{\textbf{#1}}
\setlength{\tabcolsep}{4pt}

\begin{table*}[t]
  \caption{\textbf{Sample Efficiency Comparison}. The number of environment steps (\emph{Step}) and failed episodes (\emph{Fail}) required to achieve target goal performance : (success rate, execution time speed up vs. pre-trained model without acceleration).
  Not available (N/A) values indicate none of the evaluation satisfied the goal during the training.
  Our method can reach to each goal within smaller amount of environment steps and also less failures across all tasks.
  }
  \centering
  \label{tab:results}
  \begin{tabular}{c | cc | cc | cc | cc | cc | cc | c}
    \toprule
      Task $\rightarrow$& \multicolumn{2}{c|}{\texttt{Lift}} & \multicolumn{4}{c|}{\texttt{Can}} & \multicolumn{4}{c|}{\texttt{Square}} &\texttt{Kitchen} \\
      Goal $\rightarrow$& \multicolumn{2}{c|}{0.99,×2} & \multicolumn{2}{c|}{0.99,×2} & \multicolumn{2}{c|}{0.99,×3} & \multicolumn{2}{c|}{0.97,×2} & \multicolumn{2}{c|}{0.97,×2.5} & 0.99,×1.5 \\
$\downarrow$ Method & Step   & Fail   & Step   & Fail   & Step   & Fail & Step   & Fail   & Step    & Fail & Step \\
    \midrule
\texttt{No-Accel}   &3.6M    &5.5K    & 6.1M   &5.8K    &\mcr{N/A}      &19M     &22K     &46M      &56K    &28M    \\
\texttt{\acronym}   &\q{0.7M}&\q{1.4K}&2.6M    &\q{3.6K}&\q{10M}&\q{14K}&\q{9.3M}& \q{16K}&\q{16M}  &\q{24K}&\q{13M}\\
\texttt{Accel(x2)}  &2.1M    &3.4K    &3.7M    &5.0K    &\mcr{N/A}      &10.9M   &19K     &25M      &37K    &22M    \\
\texttt{Accel(x3)}  &0.9M    &3.1K    &\q{2.1M}&5.6K    &18M    &28K    &31M     &79K     &31M      &79K    &N/A    \\
\texttt{SpeedTuning}&\mcr{N/A}        & \mcr{N/A}       &\mcr{N/A}      &\mcr{N/A}        &\mcr{N/A}        &N/A    \\
\texttt{DemoSpeedup}&3.9M    &5.8K    &4.2M    &5.6K    &\mcr{N/A}      &12M     &23K     &31M      &48K    &20M    \\
    \bottomrule
  \end{tabular}
  \label{tab:main_table}
  \vspace{-0.5cm}
\end{table*}

\begin{figure*}[t]
  \centering
  \includegraphics[width=\linewidth]{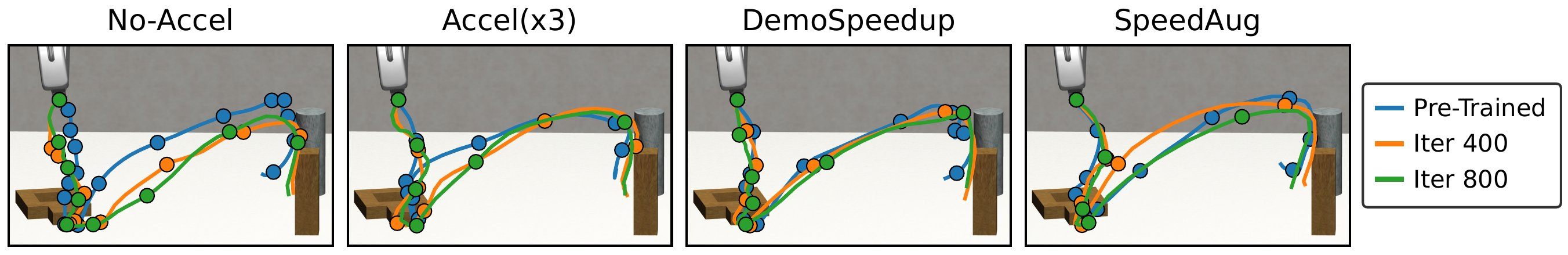}

  \caption{
  \textbf{Trajectory Visualization.} End-effector trajectories over learning progress are visualized in the \texttt{Square} task space. Each circle represents 8 environment steps.
  \texttt{SpeedAug} empirically selects fastest feasible tempos and fine-tunes the accelerated actions, which prevents learning inefficient winding trajectories by choosing overly high tempos.
  }
  \label{fig:analysis_square}
\vspace{-0.5cm}
\end{figure*}

\subsection{Simulated Manipulation Benchmarks}\label{sec:main_result}

\subsubsection{Experimental Setup}
\textbf{Tasks.}
We validate our method on three robotic manipulation tasks from the Robosuite benchmark~\citep{zhu2020robosuite} - \textit{Block Lifting} (\texttt{Lift}), \textit{Pick-and-Place Can} (\texttt{Can}), and \textit{Nut Assembly Square} (\texttt{Square}) - and the Franka Kitchen (\texttt{Kitchen}) environment~\citep{fu2020d4rl}.
The Robosuite tasks evaluate policy acceleration performance under varying task complexity and an end-effector pose action space, whereas the Kitchen task involves more complex task structures and joint-based control.
Further details are provided in~\Cref{app:tasks}.
Our main results use state-based observations (robot proprioception and object poses), while vision-based results for Robosuite tasks are reported in~\Cref{app:robosuite_img}.


\textbf{Metrics.}
To evaluate policy performance, we collect 400 rollouts and report two metrics: (1) success rate - the number of successful episodes out of 400 trials, and (2) execution time - the average length of the successful rollouts, measured in environment steps.

\textbf{Training.}
Policies are pre-trained on demonstrations using Diffusion Policy~\citep{chi2024diffusionpolicy}.
For the Robosuite tasks, we used 200 Proficient-Human demonstrations from the Robomimic dataset~\citep{robomimic2021} to pre-train the policies.
For the Kitchen environment, we used 556 demonstrations of completing four subtasks in an arbitrary order.
DPPO~\citep{dppo2024} is mainly used for RL fine-tuning in~\Cref{sec:sim_result}, 
as it stably achieves a high success rate via on-policy updates. 
PA-RL~\citep{parl} results are additionally provided in~\Cref{app:rl_alg} 
to compare sample efficiency when using an off-policy RL algorithm.
Results are reported over 4 random seeds (2 fine-tuning × 2 pre-training seeds) 
with 95\% confidence intervals.
See~\Cref{app:pretrain_detail,app:ft_detail} for further details.

\subsubsection{Results}\label{sec:sim_result}
We evaluate the effectiveness of our method in robotic manipulation tasks, as shown in~\Cref{fig:main_result}.
\texttt{\acronym} converges to faster execution time with high success rate after the initial phase of fine-tuning.
Consequently, our method significantly outperforms the baselines in sample efficiency, as summarized in~\Cref{tab:main_table}.
We report the number of environment interactions required to reach accelerated execution time goals while achieving a high success rate, in order to fairly evaluate sample efficiency under the success rate-execution time trade-off.
More results for extended target performance are provided in ~\Cref{app:robosuite_ext}.

\textbf{Robosuite Tasks.} \texttt{\acronym} required only 37\% of the environment steps on average to reach twice task execution speed compared to the random exploration baseline (\texttt{No-Accel}), while meeting the target success rate.
Meanwhile, the strongest baseline, \texttt{Accel(x3)} required 74\% of environment steps, which corresponds to 81\% more samples on average compared to \texttt{\acronym}.
Notably, the sample efficiency gain becomes larger as task complexity increases, highlighting the importance of efficient fine-tuning in complex manipulation tasks.
See~\Cref{app:robosuite_img} for vision-based experiment results.

\textbf{\texttt{Kitchen} Task.} For the \texttt{Kitchen} task, which involves 7 subtasks, \texttt{\acronym} completes up to 6 subtasks with substantially faster execution speed than other methods (results in~\Cref{app:kitchen_ext}).
In terms of success rate and execution time for 4 subtasks,  to allow easier comparison with all baselines, our model uses only 65\% of online samples to achieve twice execution speed and 99\% success rate compared to the strongest baseline, \texttt{DemoSpeedup}.
Taken together, these results confirm that \texttt{\acronym} is effective across a range of manipulation complexities.

\begin{wrapfigure}[14]{r}{0.46\textwidth}
\vspace{-0.58cm}
    \centering
    \includegraphics[width=1\linewidth]{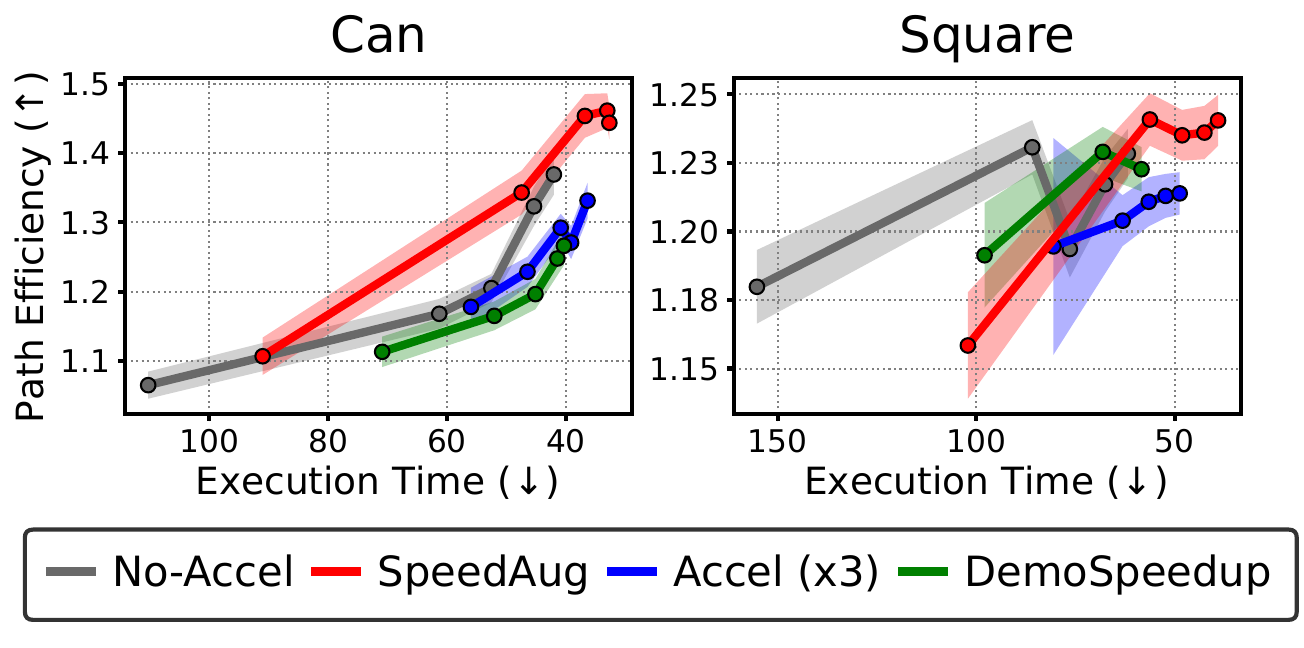}
    \caption{\textbf{Path Efficiency.} Path efficiency and execution time over RL fine-tuning progress. Path efficiency is measured by a reciprocal of the end-effector trajectory length. \texttt{SpeedAug} consistently converges to efficient trajectory by choosing feasible tempos.}
\label{fig:analysis_metric}
\end{wrapfigure}




\textbf{Path Efficiency.}
We visualize trajectory examples of each method over training progress 
in~\Cref{fig:analysis_square} to inspect path quality and performance gains over baselines.
Baselines either inefficiently improve speed by starting from the original 
demonstration tempo (\texttt{No-Accel}, \texttt{DemoSpeedup}), or learn unstable 
and winding trajectories to compensate for overspeeding (\texttt{Accel}).
In contrast, \texttt{\acronym} efficiently explores diverse tempos and converges 
to an appropriate tempo for each task segment, yielding stable and smooth trajectories.
Path efficiency comparison in~\Cref{fig:analysis_metric} further supports this, 
showing that \texttt{\acronym} converges to more efficient paths, while \texttt{Accel} 
finds substantially longer trajectories at the same execution time level.

\textbf{Failure Efficiency.} Additionally, we report the number of failed episodes needed to reach the performance goal in~\Cref{tab:main_table}.
In a real-world scenario, failed episodes can result in higher costs and safety issues compared with successful episodes.
While our method starts from a low success rate, the results demonstrate that fine-tuning quickly recovers the success rate and exhibits improved failure efficiency compared to other methods.

\subsection{Real Robot Experiments}\label{sec:robot_exp}

\subsubsection{Experimental Setup}
\begin{figure}[t]
    \centering
    \begin{subfigure}[t]{0.48\linewidth}
        \centering
        \includegraphics[width=0.99\linewidth]{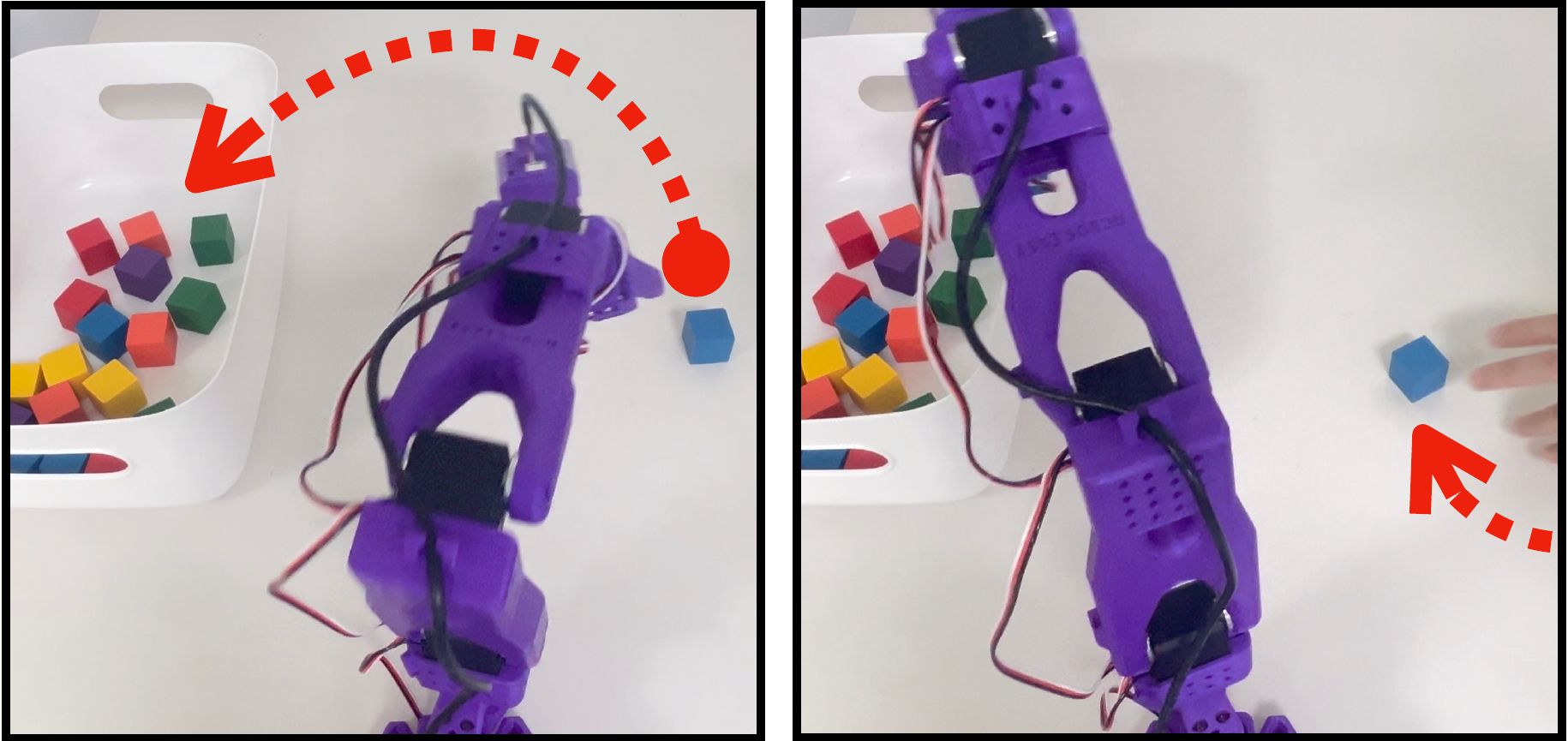}
        \caption{Continual Pick-and-Place Task}
        \label{fig:robot_task}
    \end{subfigure}
    \begin{subfigure}[t]{0.48\linewidth}
        \centering
        \includegraphics[width=0.99\linewidth]{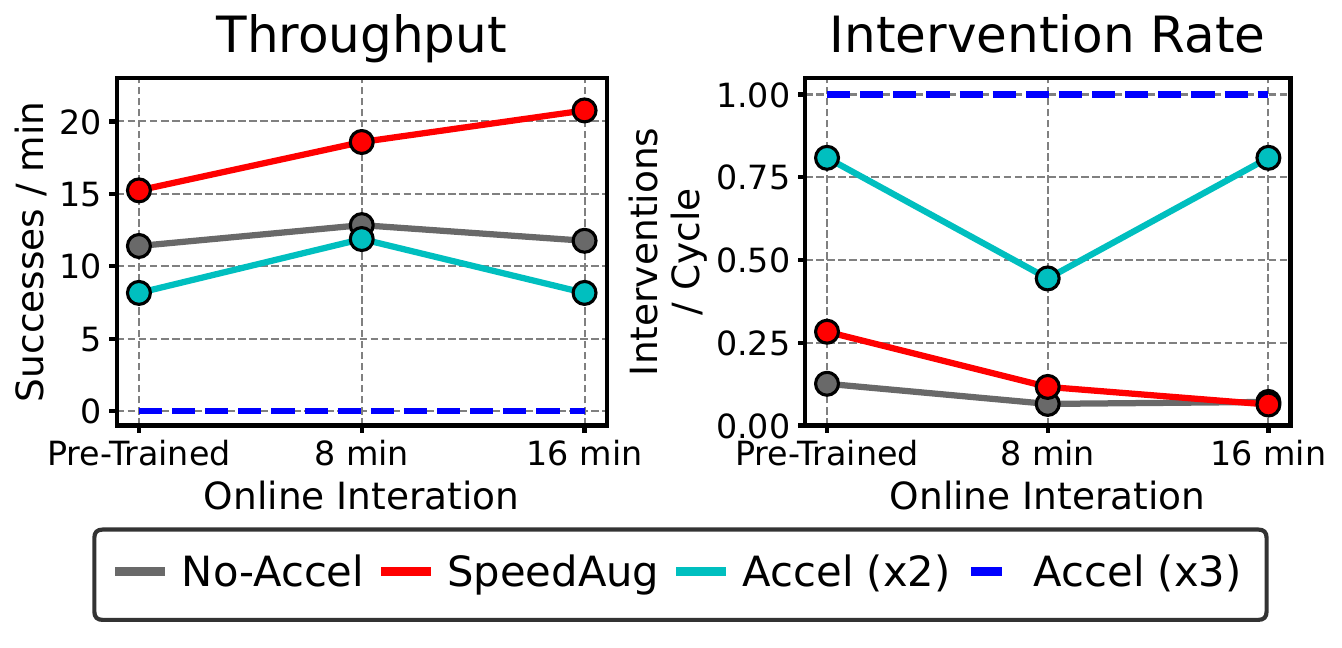}
        \caption{Online RL Results}
    \label{fig:robot_result}
    \end{subfigure}

    \caption{\textbf{Real Robot Task and Results.} Applied to a real-world manipulation task, 
    \texttt{\acronym} significantly improves task throughput using only 16 minutes of 
    environment interaction, while maintaining a high success rate.}
\label{fig:robot}
\vspace{-0.32cm}
\end{figure}

\textbf{Continual Pick-and-Place.}
We further validate our method for policy acceleration in a real robot task.
We set up a task in which the robot repeatedly picks and places blocks into a box, 
as illustrated in~\Cref{fig:robot_task}.
The blocks are continually provided by a human operator upon successful grasp.
The human operator also provides a binary reward signal upon successful placement 
and a termination signal on any unrecoverable or dangerous behavior.
The robot policy takes two camera streams (wrist and shoulder views) and controls 
the 5-DoF arm with a 1-DoF gripper using 20Hz joint PID control.
For evaluation, we execute the policy for 8 minutes in the continual task and measure 
(1) task throughput - the average number of successes per minute, and 
(2) intervention rate - the average number of system resets needed per success.
Further details are provided in~\Cref{app:tasks}.

\begin{figure*}[t]
    \centering
    \includegraphics[width=\linewidth]{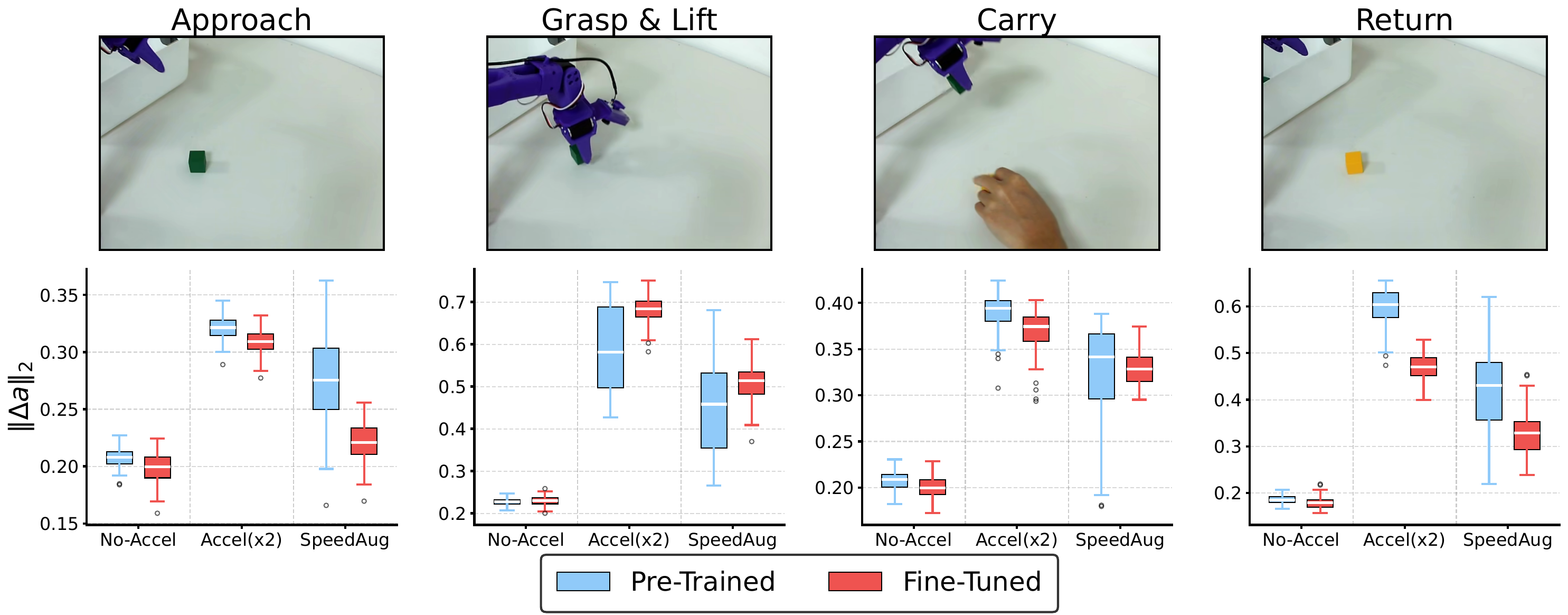}
    \caption{\textbf{Tempo Selection Results.} Action chunk magnitudes before and after RL fine-tuning 
    show that \texttt{\acronym} converges to task-appropriate tempos, while baselines struggle to explore out-of-distribution speeds.}
\vspace{-0.53cm}
\label{fig:robot_qual}
\end{figure*}

\textbf{Training.}
The policy is pre-trained on 1.4 hours of demonstrations collected by a human operator 
using a leader robot.
We used a flow matching policy and transformer-based network architecture, 
following~\citet{bjorck2025gr00t}.
Train-time RTC~\citep{traintimertc} is applied for asynchronous policy inference to remove the inconsistency between action chunks.
We fine-tuned the policy using RL over two iterations of real-world interaction.
PA-RL is used for RL updates, with a slight modification to optimize the policy within a few RL iterations.
Details are provided in \Cref{app:pretrain_detail,app:ft_detail}.

\subsubsection{Results}
\Cref{fig:robot_result} shows the online RL fine-tuning results, with additional real-robot execution examples provided in the supplementary videos.
Compared to the pre-trained policy trained on the original demonstration speed, 
\texttt{\acronym} improves the throughput by 88\% (11.4 $\rightarrow$ 20.76) 
using only 16 minutes of online interaction, without compromising the intervention rate.
While other baselines fail to learn faster or more precise task execution within the same amount of experience, our method efficiently explores and selects appropriate tempos for each task phase.
\Cref{fig:robot_qual} qualitatively supports this by comparing the magnitudes of action chunk samples before and after RL fine-tuning.
\texttt{\acronym} initially covers a range of tempos and converges to a reasonable tempo depending on the phase of the task (e.g., lifting and carrying quickly while approaching carefully for stable grasping).
In contrast, baselines struggle to obtain out-of-distribution tempos within 2 iterations 
of RL, resulting in incremental performance gains from fine-tuning.
Therefore, these results emphasize that exploring diverse tempos is important for policy acceleration in a real robot task, and \texttt{\acronym} effectively achieves it.

\begin{figure}[t]
    \centering
    
    \begin{subfigure}[t]{0.49\linewidth}
        \centering
        \includegraphics[width=1\linewidth]{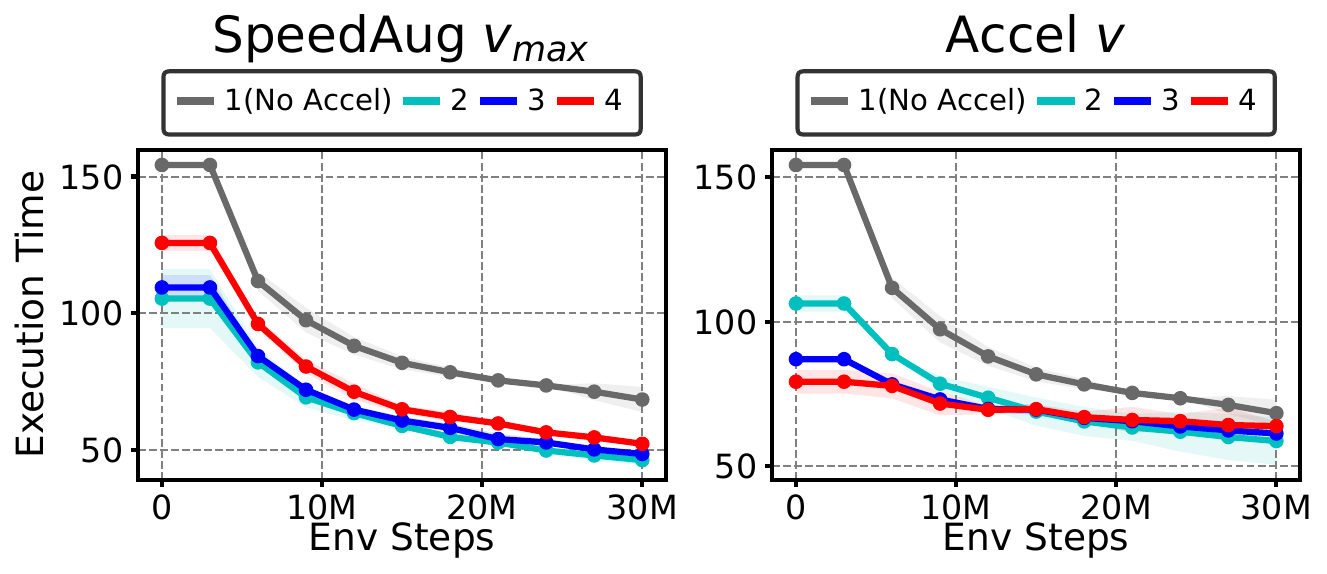}
        \caption{Sample Efficiency}
        \label{fig:analysis_metric_can}
    \end{subfigure}
    \begin{subfigure}[t]{0.49\linewidth}
        \centering
        \includegraphics[width=1\linewidth]{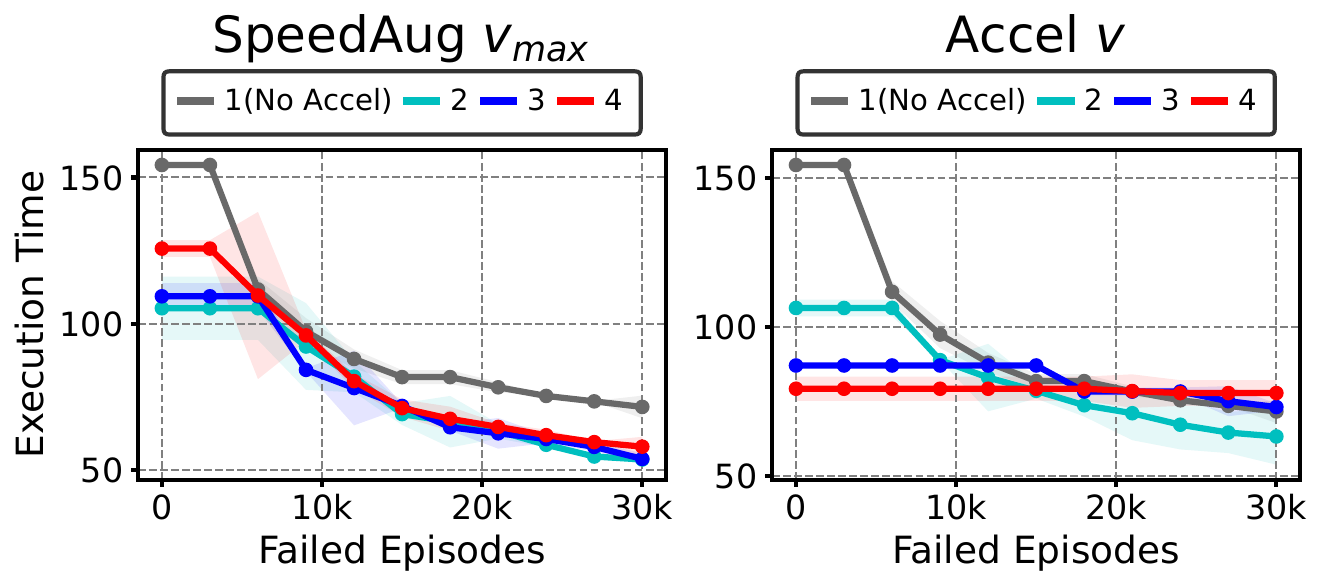}
        \caption{Failure Efficiency}
        \label{fig:analysis_metric_square}
    \end{subfigure}

    \caption{\textbf{Effect of Pre-Training Tempo.} Our method quickly selects feasible action tempos and discards overly accelerated actions. As a result, it avoids sensitivity to hyperparameter choices.}
\vspace{-0.5cm}
\label{fig:ablation}
\end{figure}

\subsection{Ablation Study: Effect of Pre-Training Tempo}\label{sec:ablation}
We further study how the choice of different tempos during pre-training affects performance in the \texttt{Square} task.
The results in~\Cref{fig:ablation} show RL fine-tuning performance under different choices of acceleration ranges or values.
\texttt{Accel} results show that increasing the pre-training tempo $v$ starts fine-tuning at a higher speed but eventually converges to same level of execution time.
However, it significantly reduces the initial success rate of the pre-trained policy, making learning efficiency worse in terms of the number of failures.
In comparison, \texttt{SpeedAug} achieves better execution times regardless of the choice of maximum tempo $v_{\text{max}}$ by quickly selecting feasible tempos during the fine-tuning phase.
Consequently, even when including the overly accelerated actions during pre-training, it can quickly discard such tempos.
Thus, our approach is inherently less sensitive to hyperparameters and more failure-efficient than other methods that use fixed tempo decisions during pre-training.

    \section{Related Work}
\textbf{Imitation Learning for Robotics.}
Recent works on Imitation Learning (IL), coupled with efficient data collection via teleoperation, have enabled training real-world robotic skills at a reasonable data collection cost~\citep{aloha,alohaunleashed,mobilealoha,chi2024diffusionpolicy}. 
Yet, policies obtained from imitation learning are often suboptimal in execution speed~\citep{yuan2025speedtuning,arachchige2025sail,park2024proleptic,guo2025demospeedup,kim2025espada} due to the suboptimality of demonstrations. 
Motivated by this observation, we aim to develop a framework that can fine-tune policies for faster task execution without requiring more demonstrations.

\textbf{RL Fine-Tuning of Generative Policy.}
As generative policies with iterative action prediction have been a common choice for policy class of pre-trained action models~\citep{chi2024diffusionpolicy,khazatsky2024droid,pi0.5,liu2024rdt,bjorck2025gr00t}, RL fine-tuning of generative policies has primarily been studied for optimizing task success rates~\citep{dppo2024,hansen2023idql,parl}.
Recent works~\citep{realtimeactionchunking,pi0.6} have reported improvements in task throughput; however, they do not explicitly explore intentional action acceleration and instead focus on developing general-purpose RL algorithms.
In this work, we use existing RL methods to accelerate pre-trained policies and improve their sample efficiency through tempo-enriched policy pre-training.

\textbf{Fast Policy Execution.}
While fast execution of robots has been a long-standing goal of robotics~\citep{nabat2005par4, constantinescu2000smooth,han2019toward,pham2019critically,li2014integrated}, recent works on robotic learning focus on enhancing success rate of challenging tasks rather than swift task execution, resulting in suboptimal execution speed of policies.
A line of work tackles this problem by skipping the first few steps of action chunks~\citep{park2024proleptic}, conditionally downsampling action chunks~\citep{yuan2025speedtuning,guo2025demospeedup,kim2025espada}, and executing action with higher control frequency~\citep{arachchige2025sail}.
Our approach aims to eliminate the a success rate trade-off introduced by heuristic acceleration methods via RL fine-tuning.
    \section{Conclusion}
We propose~\textbf{\texttt{\acronym}}, an RL-based policy acceleration framework that efficiently fine-tunes pre-trained policies for faster task execution. 
\texttt{\acronym} pre-trains a policy on speed-augmented demonstrations and constructs a behavior prior that captures diverse action tempos.
Fine-tuning the tempo-enriched policy using RL significantly improves the sample efficiency of existing RL and policy acceleration baselines without success rate trade-off.
We validated our method in simulated and real-robot manipulation tasks, 
demonstrating that our method can be practically used to obtain a high-throughput 
policy from slow demonstrations.

Despite these gains, our approach still faces several practical challenges.
Our current formulation assumes a position-based action space, which makes it difficult to directly apply our method to other common control schemes, such as velocity-based control.
Also, the initial behavior of the tempo-enriched policy stochastically executes diverse tempos, which may raise safety concerns during early-stage fine-tuning.
For future work, we expect that extending our method to a wider range of action spaces, such as relative action spaces, and combining it with heuristic tempo selection methods can further enhance the practicality of our method.



\bibliography{main}
\bibliographystyle{main}


\appendix

\newcounter{appsection}
\renewcommand{\theappsection}{\Alph{appsection}}

\newcommand{\appsection}[1]{%
  \refstepcounter{appsection}
  \section*{\theappsection\quad #1}
  \addcontentsline{toc}{section}{App.~\theappsection\quad #1}
}

\section*{Appendix}

\appsection{Societal Impact}
The motivation of this work is to improve the execution speed of robotic control, which could have diverse societal impacts.
However, we do not anticipate immediate societal impacts, as our implementation and experiments are conducted in laboratory settings and cannot be directly used for practical deployment without significant engineering effort.

\appsection{Synthetic Task Experiments : 2D Navigation}\label{app:synthetic}
\begin{wrapfigure}{r}{0.45\textwidth}
  \centering
  \includegraphics[width=\linewidth]{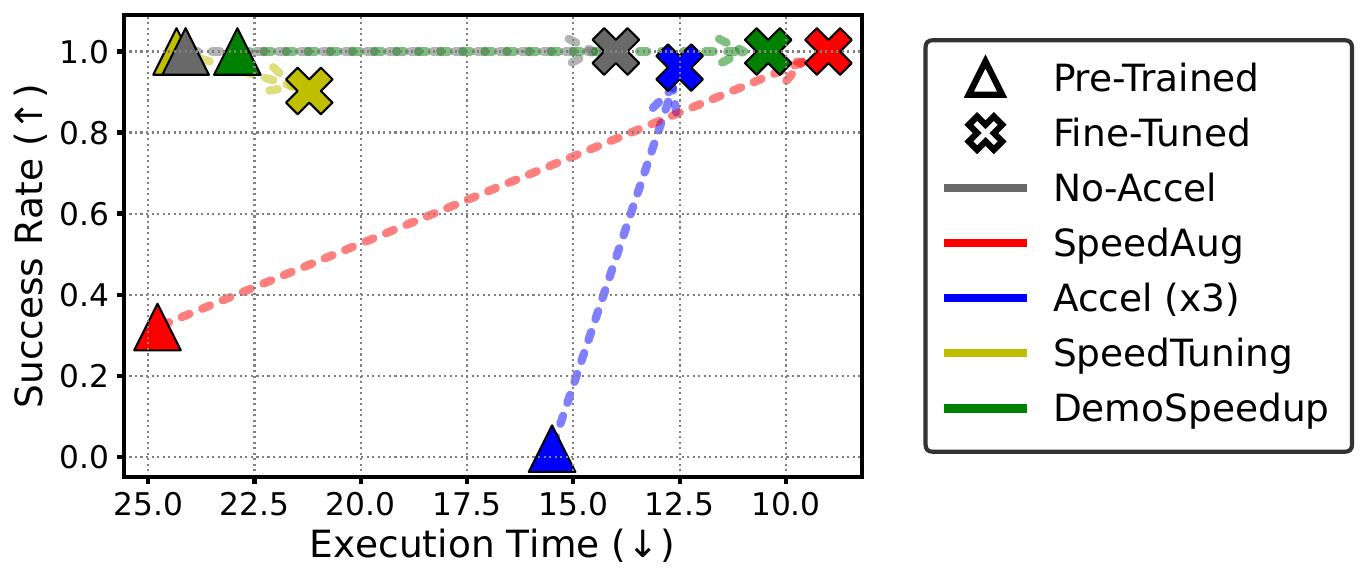}

  \caption{
  \textbf{Maze Navigation Performance.} Success rate versus execution time before ($\blacktriangle$) and after ($\boldsymbol{\times}$) the fine-tuning are shown.
  \texttt{SpeedAug} initially starts from low success rate and slow execution time exploring diverse tempos but quickly adapt to fast behavior with high success rate.
  }
  \label{fig:maze_graph}
\end{wrapfigure}
We validate our method on a simple 2D maze environment to provide intuitive understanding of the effect of our method and other baselines.
The point robot in the maze navigates by selecting the next absolute position as its action.
The maximum distance that the robot can move in a single step is limited (roughly half of the maze width).
If the robot attempts to pass through a wall, the wall reflects the trajectory with damping, which slows down the navigation.
Demonstrations are collected with restricted extension step size of RRT* planner to simulate the suboptimal speed.
Accordingly, the learning goal is to learn accelerated actions with the largest step size while avoiding collisions at corners.

\begin{figure*}[t]
  \centering
  \includegraphics[width=\linewidth]{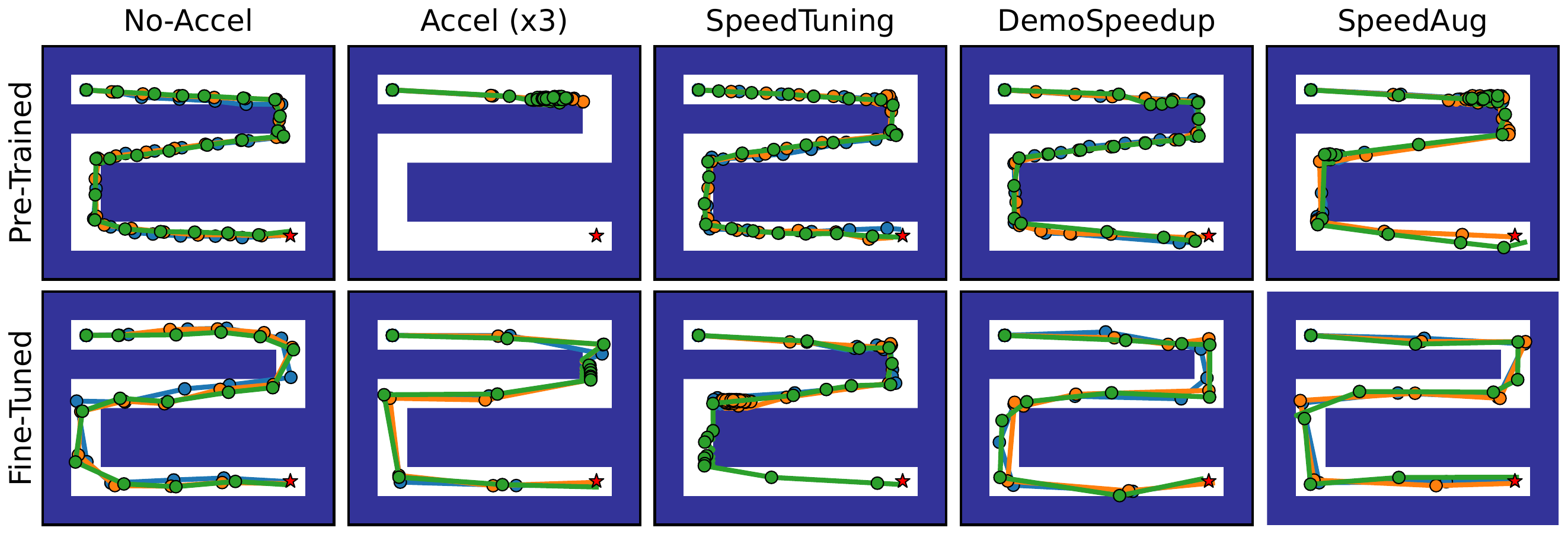}

  \caption{
  \textbf{Maze Navigation.} Three rollouts from 
  before and after the fine-tuning are visualized.
  Each circle represents single environment step. \texttt{\acronym} starts from diverse action tempos and adapts to feasible tempos with action fine-tuning, resulting in faster behavior within same number of online interactions.
  }
  \label{fig:analysis}
\end{figure*}

\begin{figure}[h]
  \centering
  \includegraphics[width=0.8\linewidth]{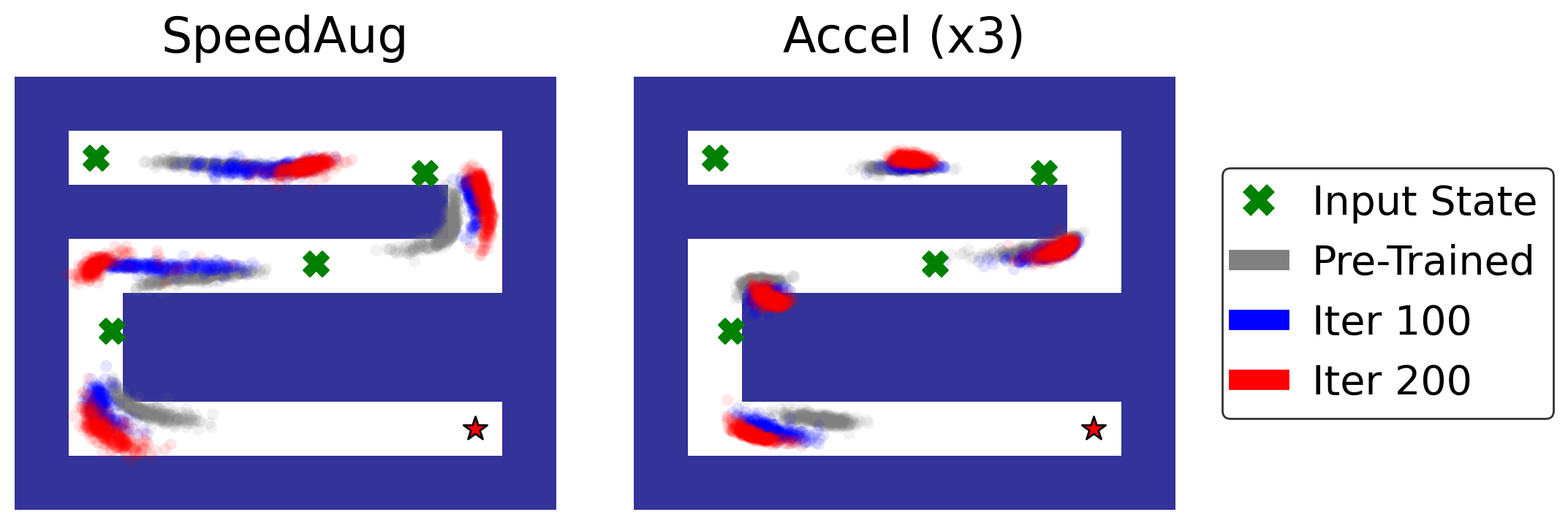}
  \caption{\textbf{Tempo Selection Visualization.}}
\label{fig:tempo_selection}
\end{figure}

Three example rollouts from policies before and after RL training are compared across different methods in~\Cref{fig:analysis}, and~\Cref{fig:tempo_selection} further visualizes the sampled actions.
As it can be seen from the results of \texttt{No-Accel}, without any intentional action acceleration, fine-tuning from the original speed of the demonstrations with random exploration is inefficient.
In contrast, \texttt{SpeedAug} initially explores diverse action tempos and quickly adapts to feasible tempos through fine-tuning.
Other baselines that start from fixed tempo selections, \texttt{Accel} and \texttt{DemoSpeedup}, effectively accelerate straight paths, but the learning progress is slower than that of \texttt{\acronym}.
\texttt{Accel} overspeeds at corners and takes a long time to escape from the locally optimal behavior of bumping into the wall.
\texttt{DemoSpeedup} successfully slows down at corners as a result of tempo selection during pre-training, but inefficiently explores faster tempos at corners, as it relies on random exploration and does not have access to different tempos of actions.

Quantitatively, the success rate and execution time are presented in~\Cref{fig:maze_graph}.
While \texttt{Accel} and \texttt{DemoSpeedup} start with faster execution times, \texttt{SpeedAug} quickly catches up in execution time and converges to a faster speed within the same number of environment steps.
On the other hand, \texttt{SpeedTuning} freezes the pre-trained policy, and thus does not improve execution time or success rate beyond the hard limit.
See~\Cref{app:maze_ext} for more discussion on baseline results.


\appsection{Off-Policy RL Results}\label{app:rl_alg}

\begin{figure}[t]
    \centering
    \includegraphics[width=0.98\textwidth]{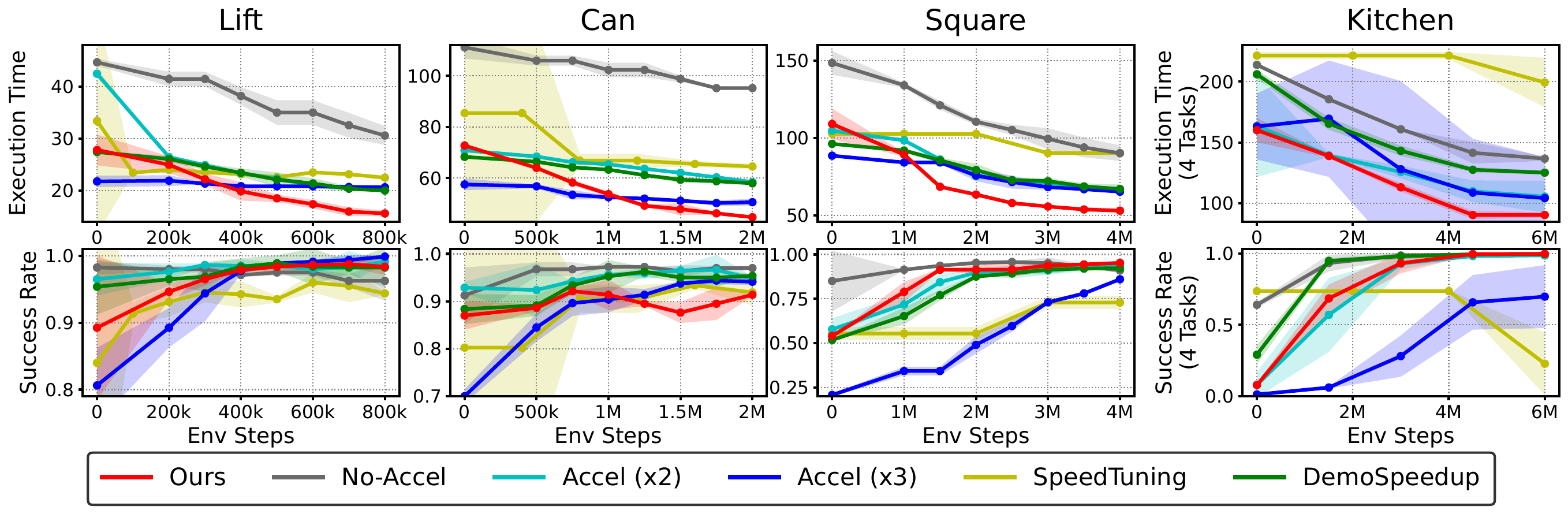}

    \caption{\textbf{Off-Policy RL Results.}}
\label{fig:alt_rl_graph}
\end{figure}

To compare sample efficiency when using an off-policy RL algorithm, we provide 
RL fine-tuning results using PA-RL~\citep{parl}.
\Cref{fig:alt_rl_graph} reports the results over 2 random seeds (1 fine-tuning × 2 pre-training).
Consistent with the results in~\Cref{fig:main_result}, \texttt{\acronym} outperforms 
the baselines in sample efficiency when using off-policy RL for fine-tuning.
PA-RL improves the overall sample efficiency across all methods, while the convergence 
gain from tempo selection of \texttt{\acronym} remains effective.
We note that, despite the improved sample efficiency of off-policy RL, it requires high updates-to-data (UTD) for critic learning, resulting in a similar level of experiment time in simulation compared to DPPO experiments.
Thus, we mainly report DPPO results, which show more stable success rate convergence, in our main text, while we use PA-RL for our real robot experiments where data collection cost is significantly higher.

\begin{figure}[h]
    \centering
    \includegraphics[width=\linewidth]{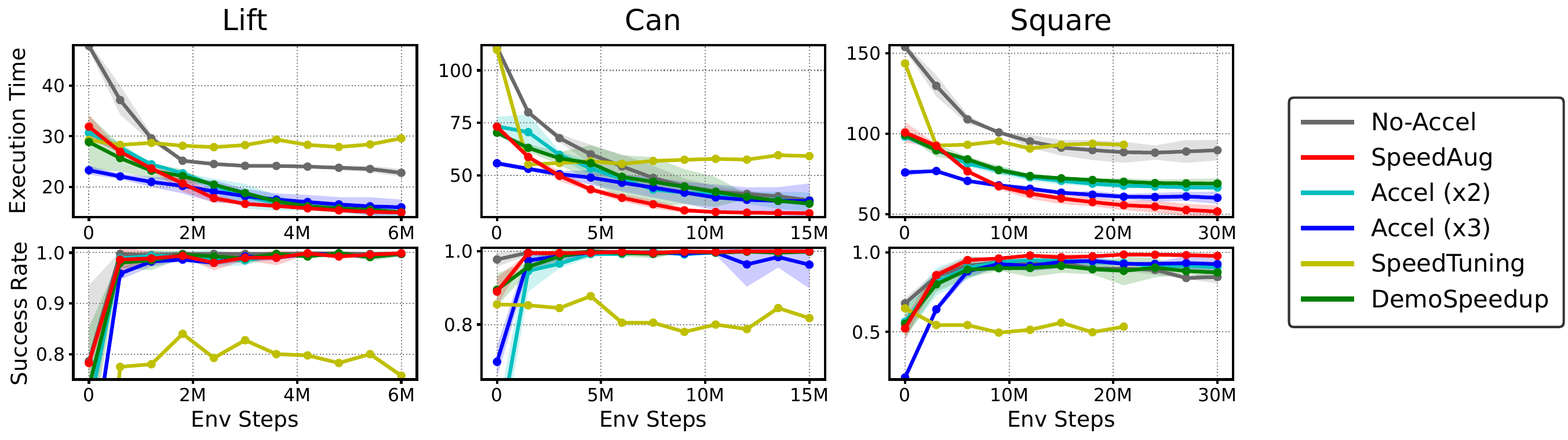}
    \caption{\textbf{Vision-based Observations Results.}}
\label{fig:robosuite_img}
\end{figure}

\appsection{Robosuite Vision-based Experiments}\label{app:robosuite_img}
Extending the state-based results, we report vision-based observation results in~\Cref{fig:robosuite_img} over 2 random seeds (1 fine-tuning × 2 pre-training).
Consistent with the results in~\Cref{fig:main_result}, our method achieves fast execution times and high success rates using fewer online samples, demonstrating its effectiveness in vision-based control settings.

\appsection{Tasks Details}\label{app:tasks}
\textbf{Robosuite.} Three tasks from Robosuite~\citep{zhu2020robosuite} use a 7D action space consisting of a 6D end-effector target pose and a 1D gripper command.
The initial positions and orientations of the objects are randomized.
The policy receives a sparse reward upon task completion.

\textbf{Franka Kitchen.} The Franka Kitchen environment~\citep{fu2020d4rl} uses a 9D action space consisting of a 7D target joint configuration and 2D gripper finger positions.
A sparse reward is given each time the agent interacts with any of the seven objects in the environment.
Thus, the goal of the policy is to interact with as many objects as possible within the time limit.

\begin{figure}[h]
  \centering
  \begin{subfigure}{0.2\linewidth}
    \centering
    \includegraphics[width=\linewidth]{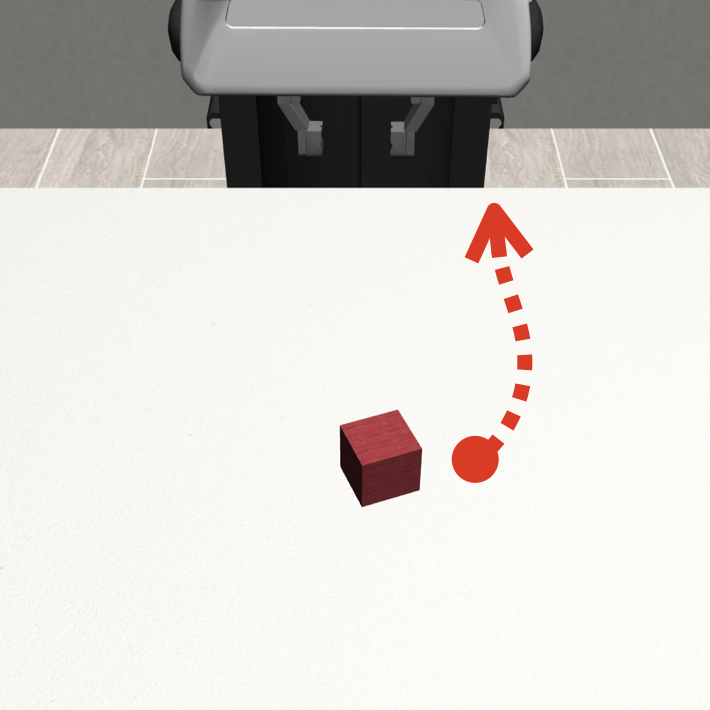}
    \caption{\texttt{Lift}}
  \end{subfigure}
  \hfill
  \begin{subfigure}{0.35\linewidth}
    \centering
    \includegraphics[width=\linewidth]{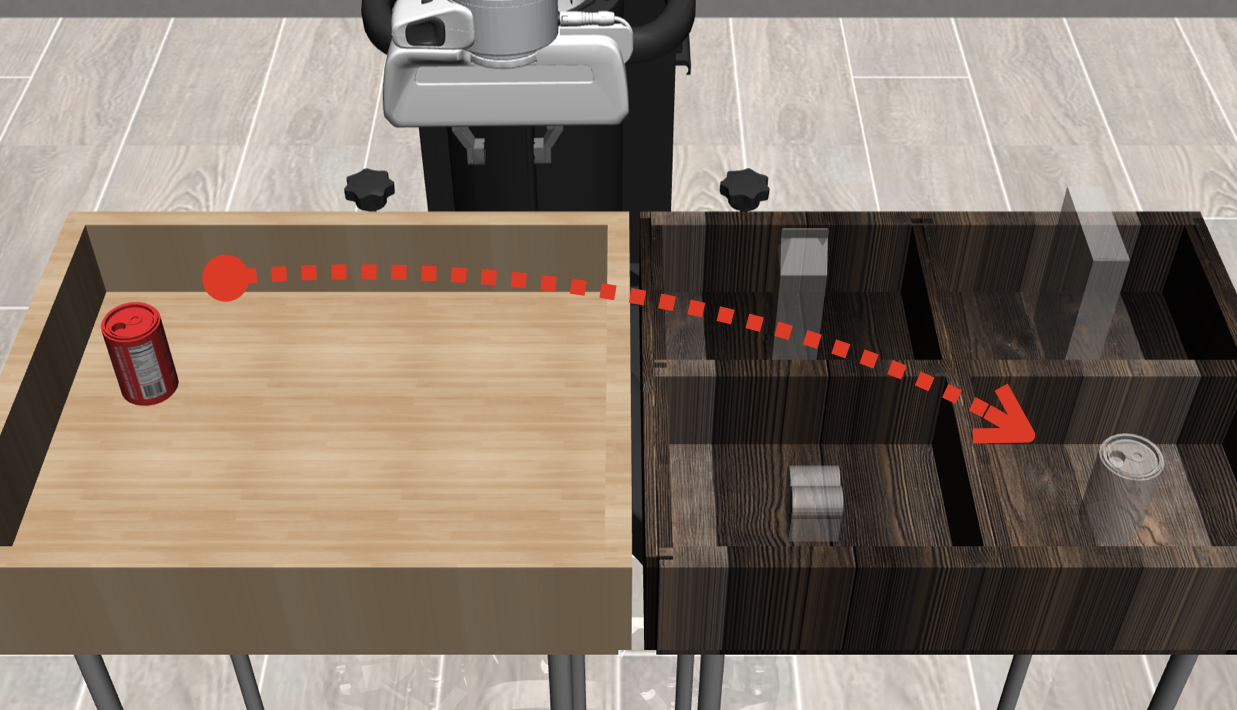}
    \caption{\texttt{Can}}
  \end{subfigure}
  \hfill
  \begin{subfigure}{0.2\linewidth}
    \centering
    \includegraphics[width=\linewidth]{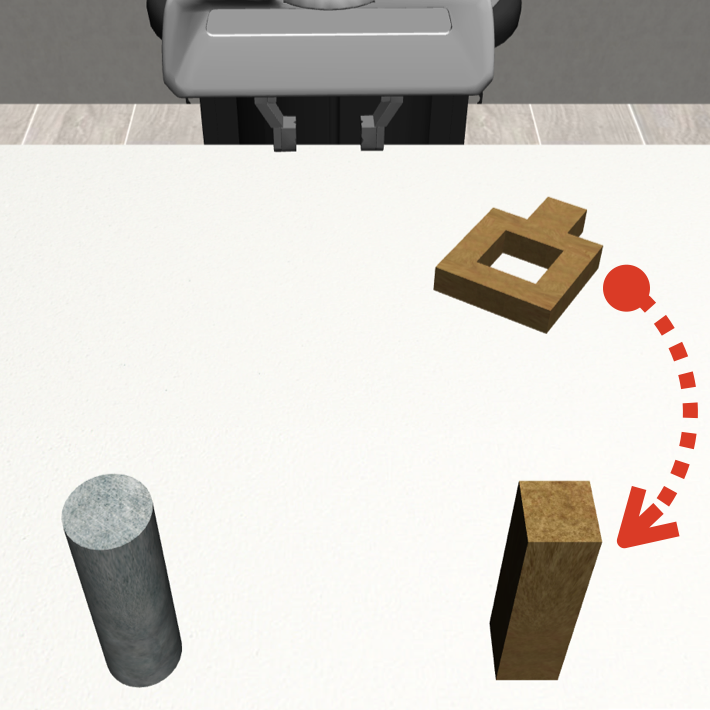}
    \caption{\texttt{Square}}
  \end{subfigure}
  \hfill
  \begin{subfigure}{0.2\linewidth}
    \centering
    \includegraphics[width=\linewidth]{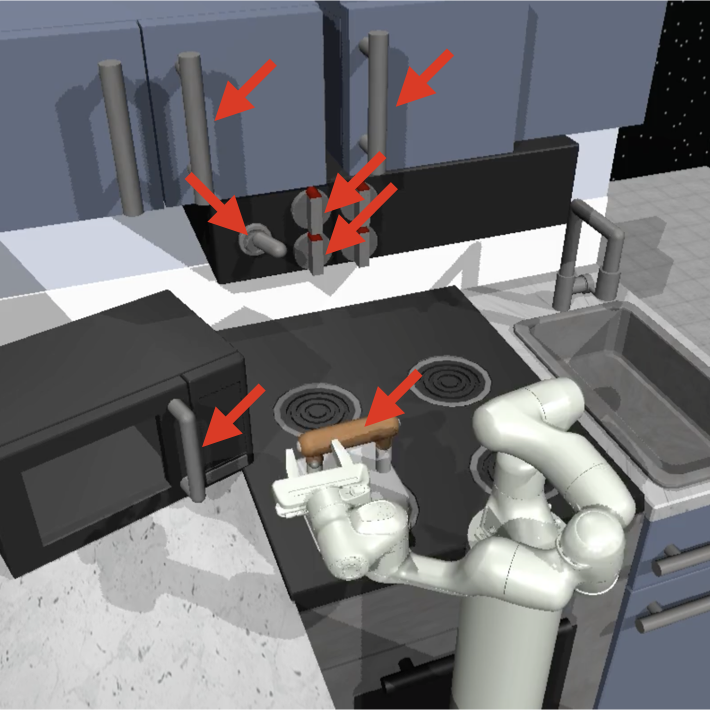}
    \caption{\texttt{Kitchen}}
  \end{subfigure}
    \caption{\textbf{Block Lifting.} Lift a block until it reaches a target height. 
\textbf{Pick-and-Place Can.}
Pick a can and place it into the correct bin. 
\textbf{Nut Assembly Square.} Fit the square-shaped nut to the matching peg. \textbf{Kitchen.} Complete 7 possible subtasks by manipulating with the objects correctly. }
  \label{fig:tasks}
\end{figure}

\begin{figure}[h]
    \centering\includegraphics[width=0.6\linewidth]{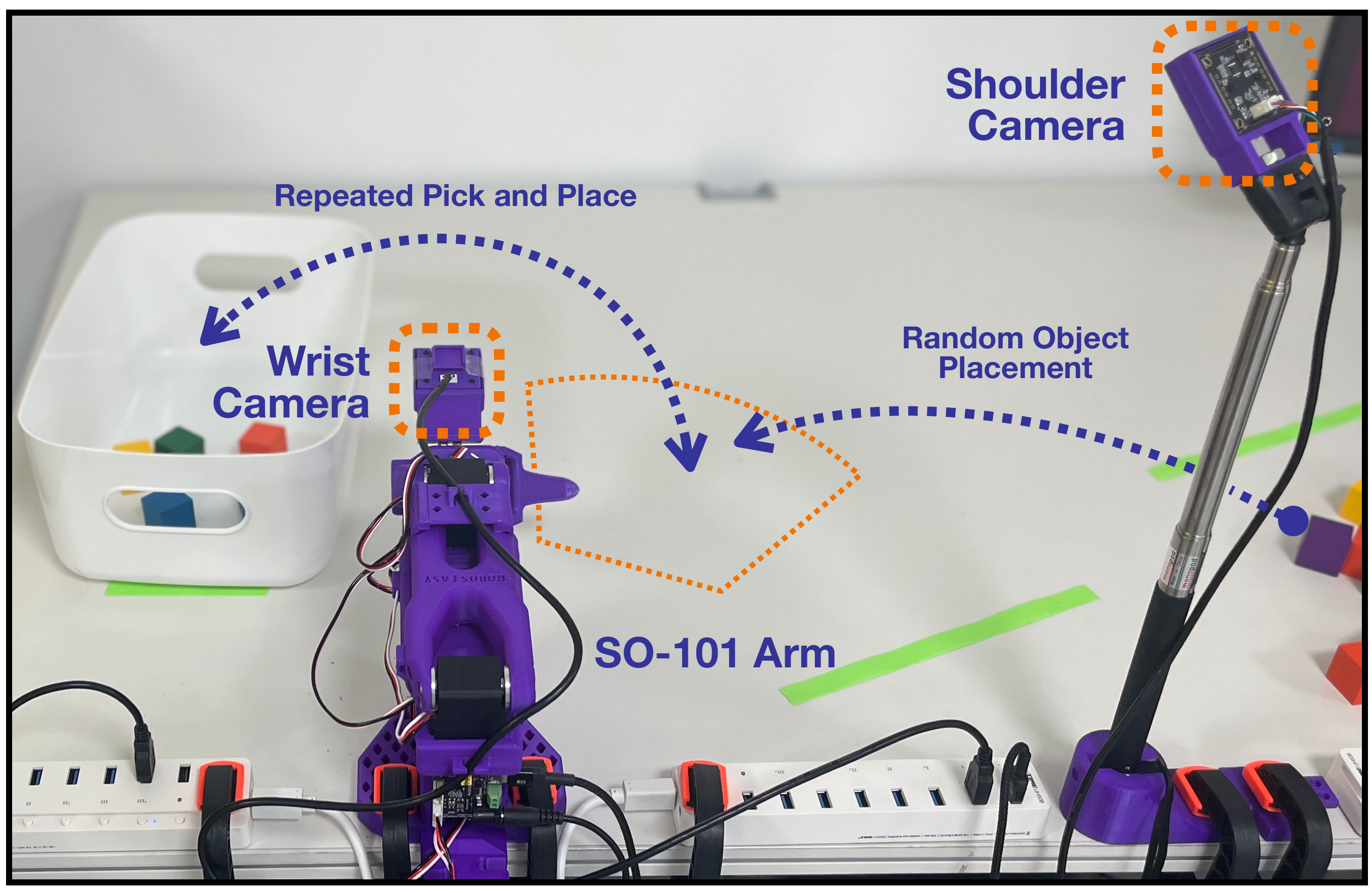}
  \caption{\textbf{Hardware and Task Space setup.}}
  \label{fig:hardware_setup}
\end{figure}

\textbf{Continual Pick-and-Place.}
We collected 1.4 hours of demonstrations containing 1000 successful pick-and-place trials over 50 episodes using a leader arm robot.
Objects are 2.5cm cubic wooden blocks in 7 different colors, placed at random reachable positions and orientations on the table.
During policy rollouts for RL fine-tuning, a human operator manually provides a binary reward signal upon successful placement of an object into a box.
Unless the operator terminates the episode, the robot continues attempting to pick and place the blocks.
Episodes are terminated when any of the following occurs:
(1) dangerous behavior is observed, such as a hard collision with the table, 
objects, or the box; or
(2) the robot enters an unrecoverable state, typically indicated by repeated 
failures to approach the object or more than three unsuccessful grasp attempts 
without improvement in grasp pose.
See~\Cref{fig:hardware_setup} for the hardware and task space setup.

\begin{figure}[h]
  \centering
  \includegraphics[width=0.4\linewidth]{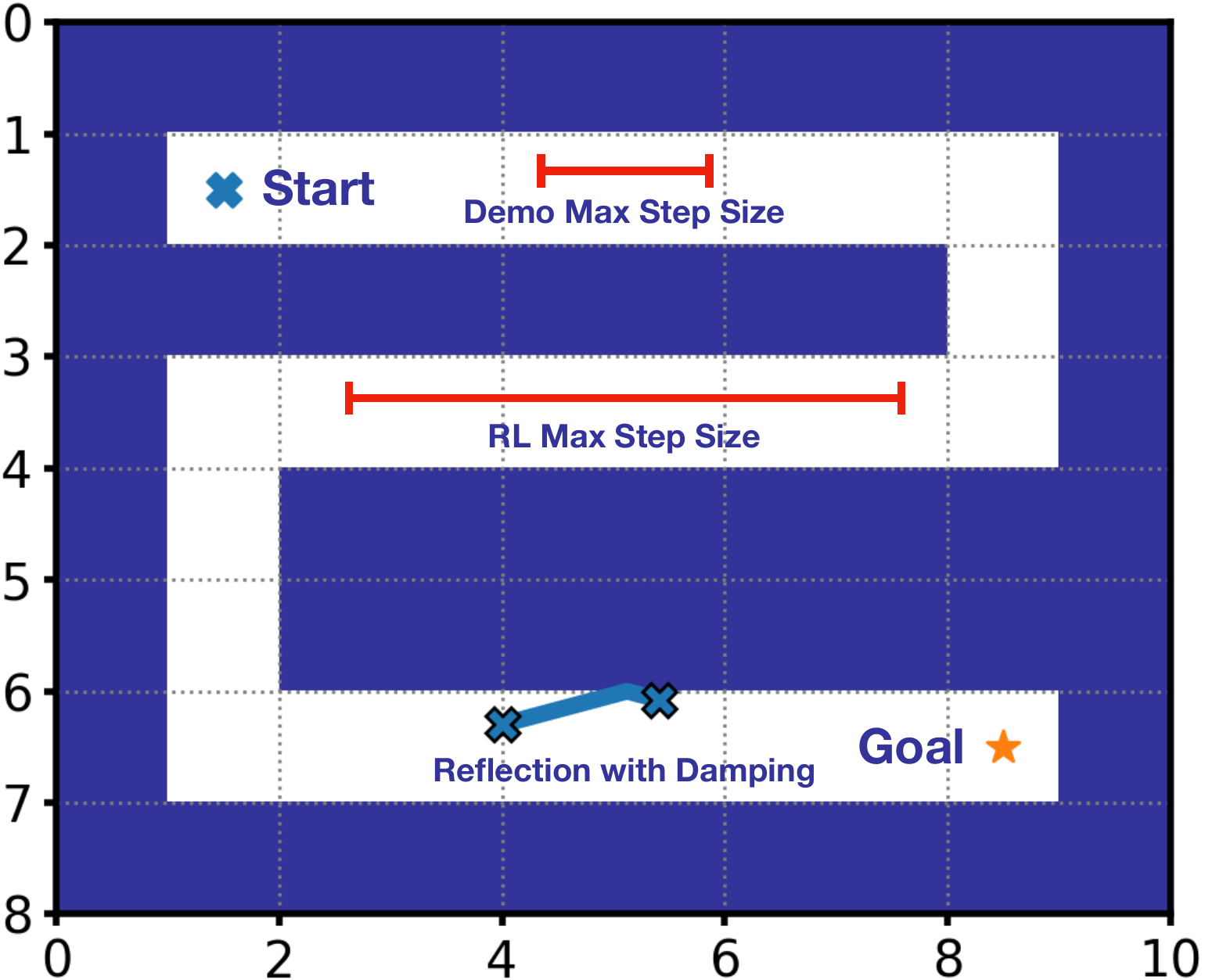}
\caption{\textbf{Maze Navigation Task.} }
  \label{fig:maze_explain}
  \vspace{-0.5cm}
\end{figure}

\textbf{Maze Navigation.} A point robot navigates toward a fixed goal by selecting the next absolute position in the maze.
The observation and action spaces are both 2D positions, and a sparse reward is provided upon task completion.
During RL fine-tuning, the robot can move up to half the maze width in a single environment step, as illustrated in~\Cref{fig:maze_explain}.
Demonstrations are collected using a smaller step size to simulate suboptimal execution speed.
Actions that result in wall collisions cause the robot to bounce according to a reflection rule, with the reflection distance capped at 0.3.

\appsection{Pre-Training Details}\label{app:pretrain_detail}
\textbf{Diffusion Policy.}
In our simulated benchmark experiments, policies are pre-trained with DDPM loss with cosine beta schedule~\citep{nichol2021improved}, following~\citep{chi2024diffusionpolicy}.
Noise predictor $\epsilon_\theta$ is U-net~\citep{unet} with two residual 1D convolution blocks for its downsampling, bottleneck, and upsampling networks.
Each residual block has two 1D convolution layers with a residual connection and also 1-layer MLP condition encoder that encodes observation and diffusion timestep.
Diffusion timestep is processed by sinusoidal embedding~\citep{vaswani2017attention} and two layers of MLPs before being fed into each residual block.

For vision-based experiments in~\Cref{fig:robosuite_img}, the image observations are embedded by ViT~\citep{vit} encoder and concatenated with the observations and timestep embedding.
For \texttt{Lift}, we use a third-person view for observation at a resolution of 96×96, whereas additional wrist camera views with the same resolution are used for \texttt{Can} and \texttt{Square}.
Hyperparameters are provided in~\Cref{tab:dp_hyper}.

\begin{table}[h]
    \caption{\textbf{Diffusion Policy Hyperparameters.}}
    \centering
    \begin{tabular}{c|c}
      \toprule
       Parameter & Value \\
      \midrule
      Time embedding dims & [16,64,16] \\
      U-net channels& [64, 128] (\texttt{Square}), [40, 80] (others) \\
      Denoising steps & 20 \\
      Action prediction horizon & \makecell{
          \texttt{SpeedTuning}: 6 (\texttt{Maze}), 32 (\texttt{others})\\
          Other methods: 4 (\texttt{Maze}), 16 (\texttt{others})
      } \\
      Action execution horizon & 2 (\texttt{Maze}), 8 (others) \\
      Learning rate & 1e-4 $\rightarrow$ 1e-5 \\
      Batch size & 256 \\
      Weight decay & 1e-6 \\
      \midrule
      ViT patch size & 8 \\
      ViT embed size & 64 \\
      ViT heads & 4 \\ 
      \bottomrule
    \end{tabular}
    \label{tab:dp_hyper}
\end{table}

\textbf{Flow Matching Policy.}
For our real robot experiments, we use a flow matching~\citep{lipman2022flow} policy, that predict action by following iterative process:

\begin{equation}
    \mathbf{a}^{(0)} \sim \mathcal{N}(0, \mathbf{I}), \;\;
    \mathbf{a}^{\tau+\Delta \tau} = \mathbf{a}^\tau + \Delta\tau \cdot v_\theta(\mathbf{s}, \mathbf{a}^\tau, \tau), \mathbf{a} = \mathbf{a}^{(1)}.
\end{equation}

Using linear interpolation target $\tilde{\mathbf{a}}^\tau=\tau \cdot \mathbf{a} + (1-\tau) \cdot \epsilon$, we trained the velocity prediction network using 

\begin{equation}
    \mathcal{L}^\text{FM} = 
    \mathbb{E}_{\mathbf{s}, \mathbf{a} \sim \mathcal{D},\tau\sim U(0, 1)} \left[
    ||v_\theta(\mathbf{s}, \tilde{\mathbf{a}}^{\tau}, \tau) - (\epsilon - \mathbf{a}) ||^2
    \right]
\end{equation}

We adopt the transformer-based flow policy architecture from~\citet{bjorck2025gr00t}, 
which uses cross-attention transformers between state-action embeddings and vision tokens.
The vision encoder is initialized with DINOv3-ViT-S+~\citep{simeoni2025dinov3} weights 
and fine-tuned during training, while transformer layers are randomly initialized.
Camera observations are resized to 256×256, the resolution on which the image encoder 
was pre-trained.
A random subset of the following image augmentations (up to 3) is applied during training:
\begin{itemize}
    \item RandomAffine (rotation: $[-5°, 5°]$, translate: $[-0.05,0.05]$)
    \item Brightness ($[0.8, 1.2]$)
    \item Contrast ($[0.8, 1.2]$)
    \item Hue ($[-0.05, 0.05]$)
    \item RGB Shuffle
    \item Saturation ($[0.5, 1.5]$)
    \item Sharpness ($[0.5, 1.5]$)
\end{itemize}

\begin{table}[h]
    \caption{\textbf{Flow-Matching Policy Hyperparameters.}}
    \centering
    \begin{tabular}{c|c}
      \toprule
       Parameter & Value \\
      \midrule
      State embedding MLP dims & [256, 256, 192] \\
      Action embedding MLP dims & [256, 256, 192] \\
      Action decoder MLP dims & [256, 256] \\
      Time embedding dims & [16,64,16] \\
      Transformer layers & 3 \\
      Transformer action heads & 6 \\
      Transformer head dim & 32 \\
      Flow inference steps & 10 \\
      Action prediction horizon & 12\\
      Action execution horizon & 4 \\
      Learning rate & 1e-4 $\rightarrow$ 0 \\
      Batch size & 240 \\
      Weight decay & 1e-6 \\
      Training-Time RTC max delay & 6 \\
      RTC action prefix length & 1 \\
      \bottomrule
    \end{tabular}
    \label{tab:fm_hyper}
\end{table}

\appsection{Fine-Tuning Details}\label{app:ft_detail}
\textbf{DPPO.}
The pre-trained policy is updated with PPO objective 
\begin{equation}
\mathbb{E}_{(\bar{s}_t, \bar{a}_t, \hat{A}_t) \sim \pi_{\text{old}}}\left[
    \min\left(
        \frac{\pi_\theta(\bar{a}_t \mid \bar{s}_t)}{\pi_{\text{old}}(\bar{a}_t \mid \bar{s}_t)} \, \hat{A}_t,\,
        \operatorname{clip}\!\left(
            \frac{\pi_\theta(\bar{a}_t \mid \bar{s}_t)}{\pi_{\text{old}}(\bar{a}_t \mid \bar{s}_t)},
            1 - \epsilon,\,
            1 + \epsilon
        \right)
        \hat{A}_t
    \right)
\right].
\end{equation}

while using $\langle  \bar{s}_{t,k}, \bar{a}_{t,k}, \bar{r}_{t,k} \rangle = \langle (s_t, a^{(k)}), a^{(k-1)}, \mathbbm{1}_{k=0} \cdot r_t \rangle$ to treat each denoising step as an MDP transition.
Generalized advantaged estimation (GAE)~\citep{gae} is used for computing advantage $\hat{A}_t$.
We fine-tuned the last 10 steps of denoising steps while we maintained the rest of denoising process by keeping the original denoising network.
Hyperparemters are provided in~\Cref{fig:dppo_hyper}.

\begin{table}[h]
    \caption{\textbf{DPPO Hyperparameters.}}
    \centering
    \begin{tabular}{c|ccccc}
      \toprule
       \multirow{2}{*}{Parameter} & \multicolumn{5}{c}{Value for Task} \\
                    &\phantom{0}\texttt{Maze}\phantom{0}& \phantom{0}\texttt{Lift}\phantom{0}  &   \phantom{0}\texttt{Can}\phantom{0}  &  \texttt{Square} & \texttt{Kitchen} \\
      \midrule
      Action execution horizon & \multicolumn{5}{c}{8} \\
      Value estimator dims & \multicolumn{5}{c}{[256,256,256]} \\
      Env steps per iteration & 900 & 7680 & 23040 & 61440 & 22400 \\
      Batch size & 450 & 480 & 1440 & 3840 & 1400 \\
      Learning rate & \multicolumn{5}{c}{1e-5} \\
      Max epochs per iteration & \multicolumn{5}{c}{10} \\
      Discount factor $\gamma$ & \multicolumn{5}{c}{0.99}  \\
      GAE $\lambda$ & \multicolumn{5}{c}{0.95} \\
      DPPO $\epsilon$ & \multicolumn{5}{c}{0.001 (k=10) $\rightarrow$ 0.01 (k=1)} \\
      DPPO discount $\gamma_\text{DENOISE}$ & \multicolumn{5}{c}{0.99} \\
      Min std $\sigma$ for denoising &\multicolumn{5}{c}{0.05}\\
      \bottomrule
    \end{tabular}
    \label{fig:dppo_hyper}
\end{table}

\textbf{PA-RL.}
PA-RL~\citep{parl} optimizes actions and distills the optimized actions into the policy through a supervised learning objective, instead of directly updating the policy network using an RL objective.
By doing so, it bypasses operations in the RL objective that may be infeasible depending on the policy class, making the algorithm policy-agnostic.

Concretely, PA-RL samples action from $\pi^\text{opt}$ instead of current policy $\pi$, where the sampling procedure of $\pi^\text{opt}$ is as follows:
\begin{align}
    &\mathcal{D}=\{ a_i\sim\pi(s, a) \}_{i=1}^{n}\quad
    \mathcal{D}^\text{Best}_k=\text{Best}_k(\mathcal{D}, \{Q(s, a_i) | a_i \in \mathcal{D}\}\quad \\
    &\mathcal{D}_k^\text{opt}=\{\text{GradientAscent}_{Q, \alpha, m}(a_i) |a_i \in \mathcal{D}^\text{Best}_k \} \\
    &\pi^\text{opt}(s|a) = \text{Cat}(\mathcal{D}_k^\text{opt}, \text{Softmax}_\tau(\{ Q(s, a^\text{opt}_i) | a_i \in \mathcal{D}_k^\text{opt}\})).
\end{align}
The procedure samples $n$ actions from $\pi$ and selects best-of-$k$ actions with the highest Q values.
Each selected action is then optimized via gradient ascent using $\nabla_a Q(s, a)$ with learning rate $\alpha$ for $m$ steps.
$\pi^\text{opt}$ is used as the behavior policy to train the Q-function in the online environment for a fixed number of steps, after which the actions sampled from $\pi^\text{opt}$ are distilled into $\pi$ for policy improvement.

We adopt PA-RL to implement actor-critic RL updates with a diffusion-based policy.
Specifically, we updated the value functions using the following objectives:
\begin{align}
\mathcal{L}_{V} &= \mathbb{E}_{(s,a)\sim \mathcal{D}} \Big[ (Q_{\hat{\theta}}(s,a) - V_\psi(s))^2 \Big] \\
\mathcal{L}_Q &= \mathbb{E}_{(s,a,r,s') \sim \mathcal{D}} \Big[ (r + \gamma V_\psi(s') - Q_\theta(s,a))^2 \Big].
\end{align}
where $\mathcal{D}$ is the online trajectory buffer.
Note that a separate value function $V_\psi$ is used to avoid computing $Q_\theta(s', a'\sim\pi)$, which would requires sampling from the diffusion-based policy.

For simulation experiments, we used diffusion step-wise behavior cloning loss for policy update inspired by DPPO:
\begin{equation}
    \mathcal{L}_\pi =  -\mathbb{E}_{\mathbf{a}^{(k)}, \mathbf{s}, \mathbf{a}^{(k-1)} \sim \widetilde{\mathcal{D}}}
    \left[
    \log \mathcal{N}(\mathbf{a}^{(k-1)}|\mu_\theta(\mathbf{s}, \mathbf{a}^{(k)}), \tilde{\beta}_k \mathbf{I} )
    \right].
\label{eq:bc}
\end{equation}
This is equivalent to a BC update in the augmented MDP of DPPO, treating every denoising step as a Gaussian policy parameterized by the denoising network.

For robot experiments, we used the original velocity prediction loss of flow-matching, combined with DPO loss~\citep{dpo} between the best and the worst action samples, to quickly optimize the policy within a limited steps of RL iterations : 

\begin{align}
    &\mathcal{L}_\pi = \mathcal{L}^\text{FM}_\pi(\mathcal{D}^\text{Best}_k) - \alpha \cdot \mathcal{L}^\text{DPO}(\mathcal{D}^\text{Best}_k, \mathcal{D}^\text{Worst}_k) \\
    &\mathcal{L}^\text{DPO} = \log \sigma \big( 
    [\mathcal{L}^\text{FM}_\pi(\mathcal{D}^\text{Best}_k) - \mathcal{L}^\text{FM}_{\pi_0}(\mathcal{D}^\text{Best}_k)] -
    [\mathcal{L}^\text{FM}_\pi(\mathcal{D}^\text{Worst}_k) - \mathcal{L}^\text{FM}_{\pi_0}(\mathcal{D}^\text{Worst}_k)]
    \big)
\end{align}

We chose the best $\alpha \in [0.01, 0.1]$ for the first iteration of RL and set to $\alpha=0$ for the second iteration. Hyperparemters are provided in~\Cref{fig:parl_hyper}.

\begin{table}[h]
    \caption{\textbf{PA-RL Hyperparameters.}}
    \label{fig:parl_hyper}
    \centering
    \begin{tabular}{c|c}
      \toprule
       Parameter & Value \\
      \midrule
      V dims & [256,256,256] \\
      Q dims & [1024, 1024, 1024] \\
      Global opt. sample size ($n$, $k$) & (5, 5) \\
      Local opt. & Disabled \\
      Policy Softmax temperature $\tau$ & Argmax (Real robot), 0.02 (\texttt{Kitchen}), 0.04 (others) \\
      Policy batch size & 768 (Real robot), 1024 (others) \\
      Policy learning rate & 1e-4 (Real robot), 1e-5 (others) \\
      Policy UTD & 1 \\
      Critic batch size & 256 \\
      Critic learning rate & 3e-4 \\
      Critic UTD & 8 \\
      \bottomrule
    \end{tabular}
\end{table}

\appsection{Baseline Details}\label{app:baseline_detail}

\begin{table}[h]
\caption{\textbf{Baseline Summary.}}
\centering
\small
\begin{tabular}{ccc}
\toprule
Method & \makecell{Tempo Selection} & Fine-Tuning \\
\midrule
\texttt{No-Accel}        & \ding{55} & \ding{51} \\
\texttt{Accel}           & Constant  & \ding{51} \\
\texttt{DemoSpeedup(+RL)}& Entropy   & \ding{51} \\
\texttt{SpeedTuning}     & RL        & \ding{55} \\
\texttt{SpeedAug (Ours)}             & RL        & \ding{51} \\
\bottomrule
\end{tabular}
\label{tab:tempo_comparison}
\end{table}

\textbf{No-Accel.}
RL-baseline that fine-tunes pre-trained model using DPPO without any modification for policy acceleration.

\textbf{\acronym (ours).}
\Cref{eq:accel} is applied during pre-training stage to train a policy on synthetically accelerated actions by random sampled acceleration factor $v \in \text{Uniform}(1, v_\text{max})$.
The results in~\Cref{fig:main_result} and~\Cref{tab:main_table} use $v_\text{max}=3$ for all tasks, while our ablation results in~\Cref{fig:ablation} studies the effect of different value.

\textbf{Accel ($v$).}
\Cref{eq:accel} is applied during pre-training stage to train a policy on synthetically accelerated actions by constant acceleration factor $v$.
The results in~\Cref{fig:main_result} and~\Cref{tab:main_table} use $v=3$ for all tasks, while our ablation results in~\Cref{fig:ablation} studies the effect of different value.


\textbf{SpeedTuning.}~\citep{yuan2025speedtuning}
A separated policy $\pi_\text{speed}$ is trained to decide a discrete choice of the speed up factor $v \in \{ v_1, \cdots, v_n \}$ given current state.
Predicted $v$ is used to accelerate predicted actions of pre-trained model using~\Cref{eq:accel}.
DQN~\citep{dqn} is used as an RL algorithm for training, along with distributional Q function and $k$ step bellman backup, following~\citep{hessel2018rainbow} for learning efficiency and stability.
$\{ v_1,\cdots, v_n \} = \{ 1, 2, 3, 4 \}$ is used for the Robosuite tasks and $\{ v_1,\cdots, v_n \} = \{ 1, 2, 3\}$ is used for the Kitchen task.
Hyperparameters are provided in~\Cref{fig:speed_hyper}.

\begin{table}[h]
    \caption{\textbf{SpeedTuning Hyperparameters.}}
    \centering
    \begin{tabular}{c|c}
      \toprule
       Parameter & Value \\
      \midrule
      Q dims & [1024,1024,1024,1024] \\
      Q number of bins & 100 \\
      Max Bellman backup step & 1 (\texttt{Maze},\texttt{Lift}), 3 (others)\\
      Batch size & 256 \\
      Learning Rate & 3e-4 $\rightarrow$ 3e-5 \\
      \bottomrule
    \end{tabular}
    \label{fig:speed_hyper}
\end{table}

\textbf{DemoSpeedup.}~\citep{guo2025demospeedup}
The action entropy of the pre-trained policy is used as a heuristic to identify low-precision and high-precision segments.
Actions in low-precision and high-precision segments are accelerated with $v_{\text{high}}$ and $v_{\text{low}}$, respectively.
The final policy is trained using the partially accelerated demonstrations.
Although \citet{guo2025demospeedup} does not propose fine-tuning the pre-trained policy, we compare our method against fine-tuned versions of \texttt{DemoSpeedup} as a baseline that relies on fixed, heuristic tempo selection.
We follow~\citet{guo2025demospeedup} for method details, including clustering and action acceleration.

\begin{table}[h]
    \caption{\textbf{DemoSpeedup Hyperparameters.}}
    \centering
    \begin{tabular}{c|c}
      \toprule
       Parameter & Value \\
      \midrule
      Minimum cluster size & 3 (\texttt{Maze}), 5  (others) \\
      ($v_\text{low}$,$v_\text{high}$) & (1, 3) (\texttt{Kitchen}), (2, 3) (others) \\
      Entropy threshold & 1\\
      \bottomrule
    \end{tabular}
    \label{fig:demospeed_hyper}
\end{table}

See~\Cref{tab:tempo_comparison} for a summary of baseline methods.

\appsection{Compute Resources}
For simulation experiments, pre-training a single policy took 3 hours (\texttt{Lift}), 
8 hours (\texttt{Can}, \texttt{Square}), and 20 hours (\texttt{Kitchen}) on a 24GB GPU 
(A5000) with a server-grade CPU (Xeon(R) Silver).
A single fine-tuning run using DPPO took 8 hours (\texttt{Lift}), 20 hours (\texttt{Can}), 
and 40 hours (\texttt{Square}, \texttt{Kitchen}), with 20 parallel simulated environments 
on the same system.

\begin{figure*}[h]
  \centering
  \includegraphics[width=0.7\linewidth]{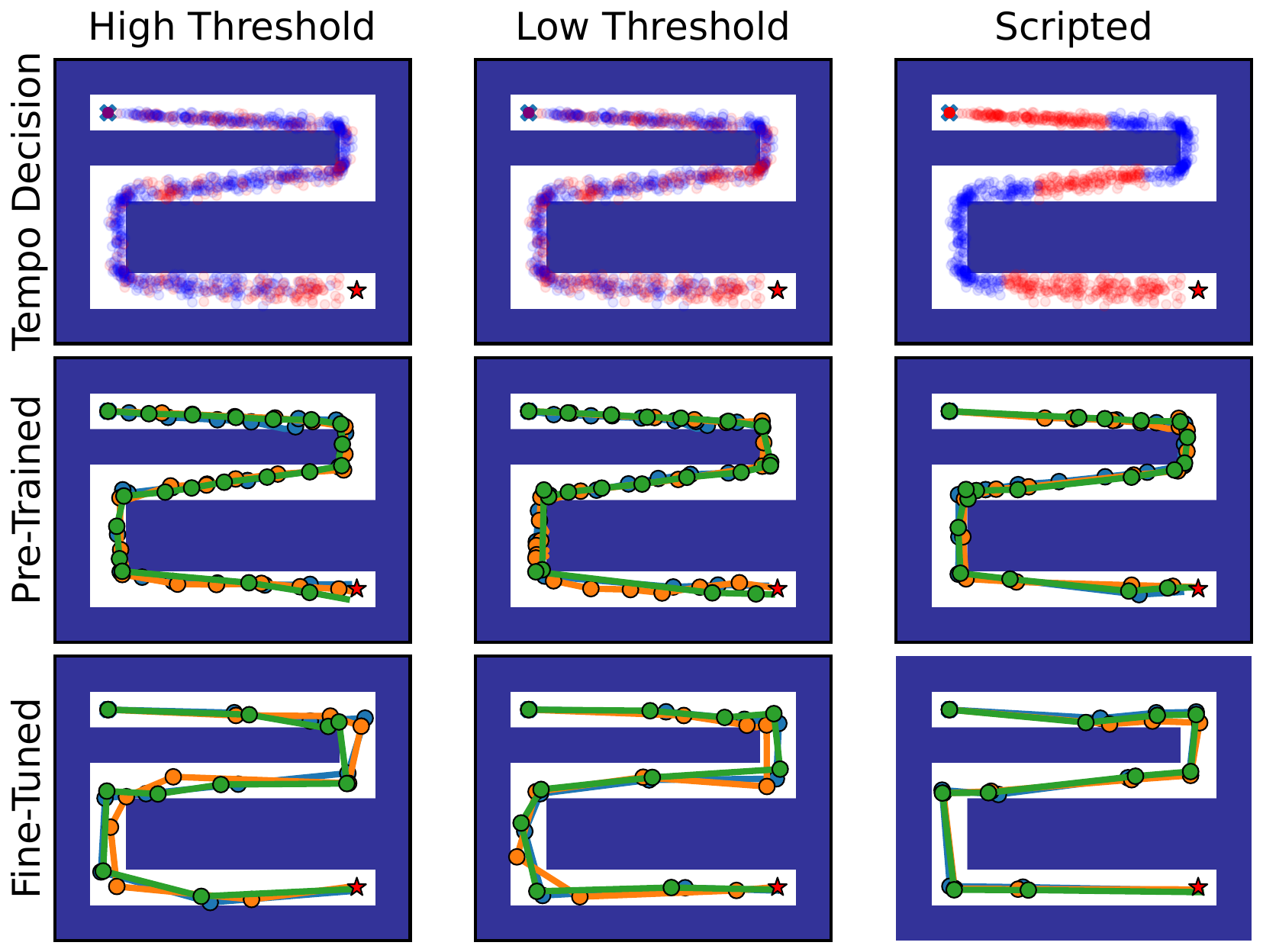}
  \caption{\textbf{\texttt{DemoSpeedup} Oracle Tempo Selections.} Top row visualizes tempo decisions used for pre-training. Red and blue circles denote acceleration with high (×3) and low (×1) tempos, respectively. Left two columns used entropy-based tempo decision and the last column used rule-based decisions.}
  \label{fig:maze_demospeed}
\end{figure*}

For real robot experiments, we used a single 48GB GPU (A6000) or two 24GB GPUs (A5000) 
for pre-training with a global batch size of 240, taking approximately 1 day.
For inference, an Alienware desktop with an RTX 4090 GPU was used, achieving 
$\sim$50\,ms total inference latency and enabling asynchronous action execution 
with only 1--2 steps of delay at 20\,Hz control.
A single iteration of RL fine-tuning took 6 hours on a single 24GB GPU (RTX 4090).
\appsection{Maze Navigation Baseline Analysis}\label{app:maze_ext}
\textbf{\texttt{DemoSpeedup}}
Since the maze navigation domain is synthetic and demonstrations are collected from a motion planner rather than a human operator, the entropy-based segmentation heuristic proposed in~\citet{guo2025demospeedup} may be less effective.
To examine this, we evaluate \texttt{DemoSpeedup} using a scripted tempo selection that approximates oracle-level decisions.
As shown in the top row of~\Cref{fig:maze_demospeed}, we design a rule that accurately distinguishes corner states from straight-path states and use it to guide acceleration.
The results in~\Cref{fig:maze_demospeed} and~\Cref{tab:demospeed_tempo} indicate that our scripted decisions do not improve the final execution time and instead slightly reduce execution speed, compared to the entropy-based tempo selection.

We identify two qualitative reasons for this behavior.
First, the boundaries of the scripted tempo decisions are not tightly localized around corners, leading to suboptimal behavior before and after corner transitions.
This highlights the sensitivity of~\citet{guo2025demospeedup} to the quality of tempo decisions and suggests the need for careful hyperparameter tuning for efficient fine-tuning.
Second, the noisy entropy estimates and the resulting segmentation expose \texttt{DemoSpeedup} to multiple action tempos, similar to \texttt{SpeedAug}, which enables efficient adaptation during the fine-tuning stage.
Overall, these observations align with the claims of this paper and support the motivation of our approach.

\begin{table}[h]
    \caption{\textbf{\texttt{DemoSpeedup} Oracle Tempo Selection Results.}}
    \label{tab:demospeed_tempo}
    \centering
    \begin{tabular}{c|c}
      \toprule
       Tempo Selection & Execution Time (steps) \\
       \midrule
       \texttt{DPPO} & 14.0\\
       \texttt{SpeedAug} & 9.0 \\
       \midrule
       Entropy (low threshold) & 10.1 \\
       Entropy (high threshold) & 10.4\\
       Scripted & 11.0\\
      \bottomrule
    \end{tabular}
\end{table}

\begin{figure}[h]
  \centering
  \includegraphics[width=0.4\linewidth]{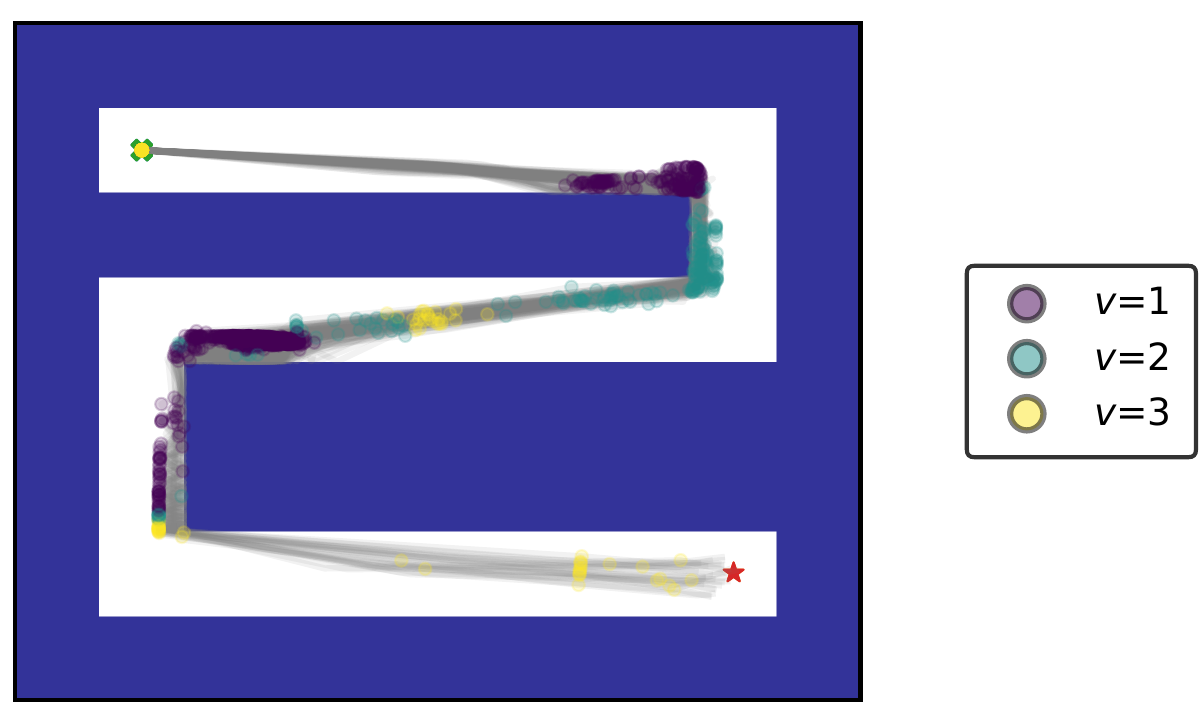}
\caption{\textbf{\texttt{SpeedTuning} Tempo Selections.} }
  \label{fig:maze_speed}
\end{figure}

\textbf{\texttt{SpeedTuning}} 
The rollouts and action tempo decisions of the \texttt{SpeedTuning} policy are shown in~\Cref{fig:maze_speed}.
The results demonstrate that \texttt{SpeedTuning} learns a meaningful policy that selects action tempos according to the maze layout - slowing down at corners and accelerating along straight paths.
However, the resulting behavior remains suboptimal, as the policy is forced to choose between faster but failure-prone tempos and slower, safer ones, without the option to fine-tune actions at the fastest successful tempo.

\begin{figure}[h]
    \centering
    \includegraphics[width=0.8\textwidth]{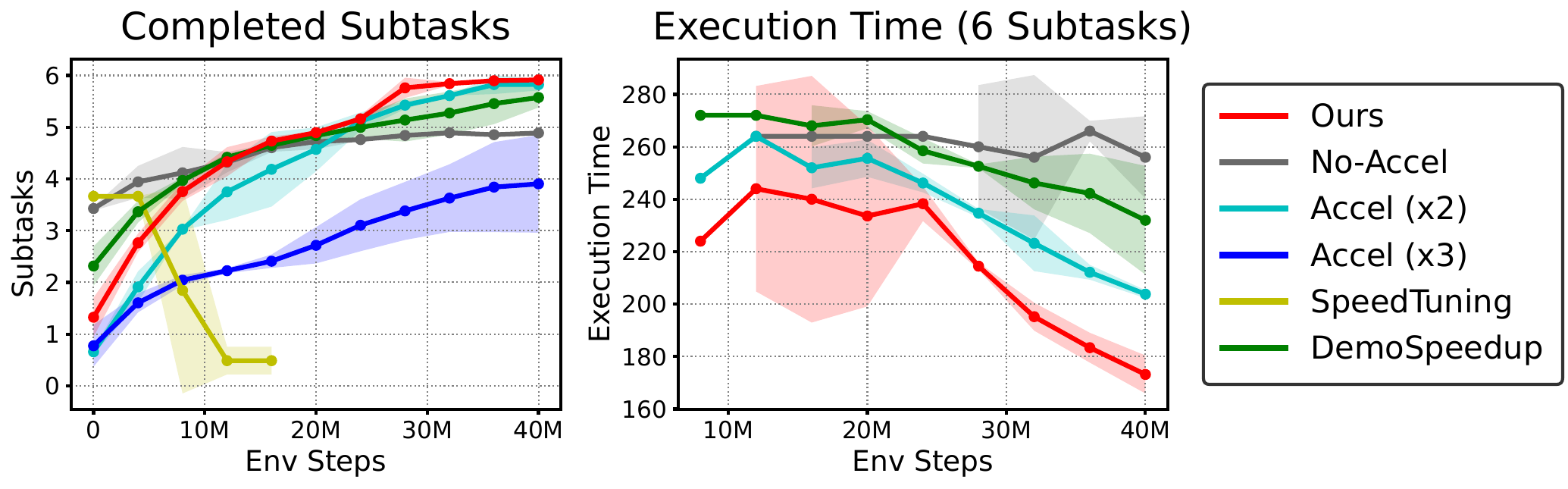}
    \caption{\textbf{\texttt{Kitchen} 6 Substasks Completion Results}}
\label{fig:kitchen_ext}
\end{figure}

\appsection{Extended Kitchen Results}\label{app:kitchen_ext}
We report the success rate and task execution time for completing 4 subtasks in our main result~\Cref{fig:main_result} for easier comparison with baselines, while there are up to 7 subtasks to complete in the \texttt{Kitchen} task.
\Cref{fig:kitchen_ext} additionally shows the average number of subtasks completed within the time limit.
Only \texttt{SpeedAug} and \texttt{Accel(x2)} reaches the average 6 subtasks completion and our method significantly outperforms the 6 subtasks execution time.

\appsection{Extended Robosuite Results}\label{app:robosuite_ext}
We provide additional sample efficiency comparison results for a wider range of target performance in~\Cref{tab:robosuite_ext}.
As observed in~\Cref{tab:main_table}, our method outperforms all baselines in terms of sample efficiency and exhibits fewer failures across all tasks.
While baseline performance overlaps with our method at modest acceleration targets ($\times$1.5), our approach remains consistently comparable to the best-performing baseline, with only marginal gaps.
We note that \texttt{SpeedTuning} particularly struggles to reach the target performance, as achieving high success rates is difficult on every tasks without fine-tuning the pre-trained policy, which supports the motivation of our method.

\setlength{\tabcolsep}{5pt}

\begin{table}[h]
  \caption{\textbf{Robosuite Extended Sample Efficiency Comparison}.
  }
  \centering
  \label{tab:results}
  \begin{tabular}{c | cc | cc | cc | cc}
    \toprule
      Task $\rightarrow$& \multicolumn{8}{c}{\texttt{Lift}} \\
      Goal $\rightarrow$& \multicolumn{2}{c|}{0.99,×1.5} & \multicolumn{2}{c|}{0.99,×2} & \multicolumn{2}{c|}{0.99,×2.5} & \multicolumn{2}{c}{0.99,×3} \\
      $\downarrow$ Method & Step & Fail & Step & Fail & Step & Fail & Step & Fail \\
    \midrule
\texttt{No-Accel}    &1.3M    &1.6K    &3.6M     &5.5K    &4.3M    &6.7K    &\mc{N/A}          \\
\texttt{\acronym}    &0.5M    &1.2K    &\q{0.7M} &\q{1.4K}&\q{1.2M}&\q{2.0K}& \q{4.4M}&\q{5.7K}\\
\texttt{Accel(x2)}   &0.9M    &1.6K    &2.1M     &3.4K    &4.2M    &6.6K    & \mc{N/A}         \\
\texttt{Accel(x3)}   &0.9M    &3.1K    &0.9M     &3.1K    &1.8M    &4.6K    &4.7M     &8.8K    \\
\texttt{SpeedTuning} &\mcr{N/A}        &\mcr{N/A}         & \mcr{N/A}       & \mc{N/A}         \\
\texttt{DemoSpeedup} &\q{0.3M}&\q{0.4K}&3.9M     &5.8K    &5.9M    &8.1K    & \mc{N/A}         \\
    \bottomrule
    \toprule
      Task $\rightarrow$& \multicolumn{8}{c}{\texttt{Can}} \\
      Goal $\rightarrow$& \multicolumn{2}{c|}{0.99,×1.5} & \multicolumn{2}{c|}{0.99,×2} & \multicolumn{2}{c|}{0.99,×2.5} & \multicolumn{2}{c}{0.99,×3} \\
      $\downarrow$ Method & Step & Fail & Step & Fail & Step & Fail & Step & Fail \\
    \midrule
\texttt{No-Accel}   &3.1M    &2.8K    &6.1M     &5.8K    &14.6M   &16.4K   & \mc{N/A}          \\
\texttt{\acronym}   &1.6M    &2.4K    &\q{2.6M} &\q{3.6K}&\q{4.8M}&\q{6.6K}& \q{10.1M}&\q{13.8K}\\
\texttt{Accel(x2)}  &\q{1.4M}&\q{2.3K}&3.7M     &5.0K    &13.4M   &13.3K   & \mc{N/A}          \\
\texttt{Accel(x3)}  &3.5M    &9.4K    &3.5M     &9.4K    &9.6M    &20.0K   &28.3M    &53.6K     \\
\texttt{SpeedTuning}&\mcr{N/A}        &\mcr{N/A}         & \mcr{N/A}       & \mc{N/A}           \\
\texttt{DemoSpeedup}&2.3M    &3.4K    &4.2M     &5.6K    &11.0M   &14.1K   & \mc{N/A}          \\
    \bottomrule
    \toprule
      Task $\rightarrow$& \multicolumn{8}{c}{\texttt{Square}} \\
      Goal $\rightarrow$& \multicolumn{2}{c|}{0.97,×1.5} & \multicolumn{2}{c|}{0.97,×2} & \multicolumn{2}{c|}{0.97,×2.5} & \multicolumn{2}{c}{0.97,×3} \\
      $\downarrow$ Method & Step & Fail & Step & Fail & Step & Fail & Step & Fail \\
    \midrule
\texttt{No-Accel}    &10.9M   &\q{13.0K}&18.7M    &21.9K    &46.4M    &56.3K    & \mc{N/A}          \\
\texttt{\acronym}    &\q{9.3M}&16.4K    &\q{9.3M} &\q{16.4K}&\q{15.5M}&\q{24.3K}& \q{24.7M}&\q{35.9K}\\
\texttt{Accel(x2)}   &\q{9.3M}&16.6K    &10.9M    &18.6K    &24.7M    &36.9K    & \mc{N/A}          \\
\texttt{Accel(x3)}   &30.9M   &79.4K    &30.9M    &79.4K    &30.9M   &79.4K    & \mc{N/A}           \\
\texttt{SpeedTuning} &\mcr{N/A}         &\mcr{N/A}          & \mcr{N/A}          & \mc{N/A}              \\
\texttt{DemoSpeedup} &12.4M   &23.3K    &12.4M    &23.3K    &30.9M    &47.9K    & \mc{N/A}          \\
    \bottomrule
  \end{tabular}
  \label{tab:robosuite_ext}
\end{table}

\clearpage



\end{document}